\documentclass[journal]{IEEEtran}
\usepackage{enumitem}
\usepackage{amsmath, amssymb,bm}
\usepackage{booktabs}
\usepackage{graphicx,epstopdf,subfigure}
\usepackage{multirow}
\usepackage{tabularx,threeparttable}
\usepackage{cite}
\usepackage{array}
\usepackage{xcolor}
\usepackage{comment}
\usepackage{mathtools}
\usepackage{bbm}
\usepackage{threeparttable}
\usepackage{bm}
\usepackage[ruled]{algorithm2e}
\usepackage{amssymb}
\usepackage{pifont}
\usepackage[colorlinks=true,citecolor=blue,urlcolor=blue]{hyperref}

\ifCLASSINFOpdf
\else
\fi
\usepackage{fancyhdr}

\newcommand{\quande}[1]{{\color{cyan}{[QUAN:#1]}}}

\newcommand{\revisemajor}[1]{\textcolor{black}{#1}}

\def\eg{\textit{e.g.}}
\def\ie{\textit{i.e.}}
\def\etal{\textit{et al.}}

\begin{document}


\title{Semi-supervised Medical Image Classification with Relation-driven  Self-ensembling Model}

\author{Quande Liu,~\IEEEmembership{Student Member,~IEEE}, Lequan Yu,~\IEEEmembership{Member,~IEEE}, Luyang Luo, \\Qi Dou,~\IEEEmembership{Member,~IEEE}, and Pheng Ann Heng,~\IEEEmembership{Senior Member,~IEEE}

\thanks{Manuscript received on 2 Dec 2019, revised on 11 Apr 2020, accepted on 13 May 2020. This work is supported by Hong Kong RGC TRS project T42-409/18-R,
the Hong Kong Innovation and Technology Commission (Project No. ITS/311/18FP, ITS/426/17FP), the National Natural Science Foundation  of China with project no. U1813204, and CUHK Direct Grant for Research. \emph{(Corresponding author: Lequan Yu.)}}
\thanks{Q. Liu, L. Yu, L. Luo, Q. Dou and P. A. Heng are with the Department of Computer Science and Engineering, The Chinese University of Hong Kong (e-mails: qdliu@cse.cuhk.edu.hk,  ylqzd2011@gmail.com, \{lyluo, qdou, pheng\}@cse.cuhk.edu.hk).}
\thanks{L. Yu is also with Department of Radiation Oncology, Stanford University, Stanford, CA 94305 USA.}
\thanks{Q. Dou is also with T Stone Robotics Institute, The Chinese University of Hong Kong.}
\thanks{P. A. Heng is also with  Guangdong-Hong Kong-Macao Joint Laboratory of Human-Machine Intelligence-Synergy Systems, Shenzhen Institutes of Advanced Technology, Chinese Academy of Sciences, Shenzhen 518055, China.}
}
\markboth{IEEE Transactions on Medical Imaging}
{Shell \MakeLowercase{\textit{et al.}}: Bare Demo of IEEEtran.cls for Journals}

\maketitle


\begin{abstract}
Training deep neural networks usually requires a large amount of labeled data to obtain good performance. However, in medical image analysis, obtaining high-quality labels for the data is laborious and expensive, as accurately annotating medical images demands expertise knowledge of the clinicians. 
In this paper, we present a novel relation-driven semi-supervised framework for medical image classification. 
It is a consistency-based method which exploits the unlabeled data by encouraging the prediction consistency of given input under perturbations, and leverages a self-ensembling model to produce high-quality consistency targets for the unlabeled data. 
Considering that human diagnosis often refers to previous analogous cases to make reliable decisions, we introduce a novel sample relation consistency (SRC) paradigm to effectively exploit unlabeled data by modeling the relationship information among different samples.
Superior to existing consistency-based methods which simply enforce consistency of individual predictions, our framework explicitly enforces the consistency of semantic relation among different samples under perturbations, \revisemajor{encouraging the model to explore extra semantic information from unlabeled data}. 
We have conducted extensive experiments to evaluate our method on two public benchmark medical image classification datasets, \ie, skin lesion diagnosis with ISIC 2018 challenge and thorax disease classification with ChestX-ray14.
Our method outperforms many state-of-the-art semi-supervised learning methods on both single-label and multi-label image classification scenarios.
\end{abstract}


\section{Introduction}
%

%

Deep learning approaches have achieved remarkable success on medical image classification, usually demanding large amount of labeled data for model learning. 
For example, Tschandl~\etal~\cite{tschandl2018the} \revisemajor{released} a large scale dataset to support algorithm development for skin lesion classification. Rajpurkar~\etal~\cite{rajpurkar2017chexnet} \revisemajor{trained} a deep model with hundreds of thousands of chest x-rays for thorax diseases diagnosis.
However, in medical imaging domain, obtaining large amount of high-quality labeled data is inevitably laborious and requires expertise medical knowledge.
Considering that unlabeled data is relatively easier to collect from clinical sites, in this paper, we aim to develop a semi-supervised medical image classification algorithm to reduce the labour of large scale data annotation, by effectively leveraging the unlabeled data.


%
\begin{figure}[t]
	\centering
	\includegraphics[width=0.47\textwidth]{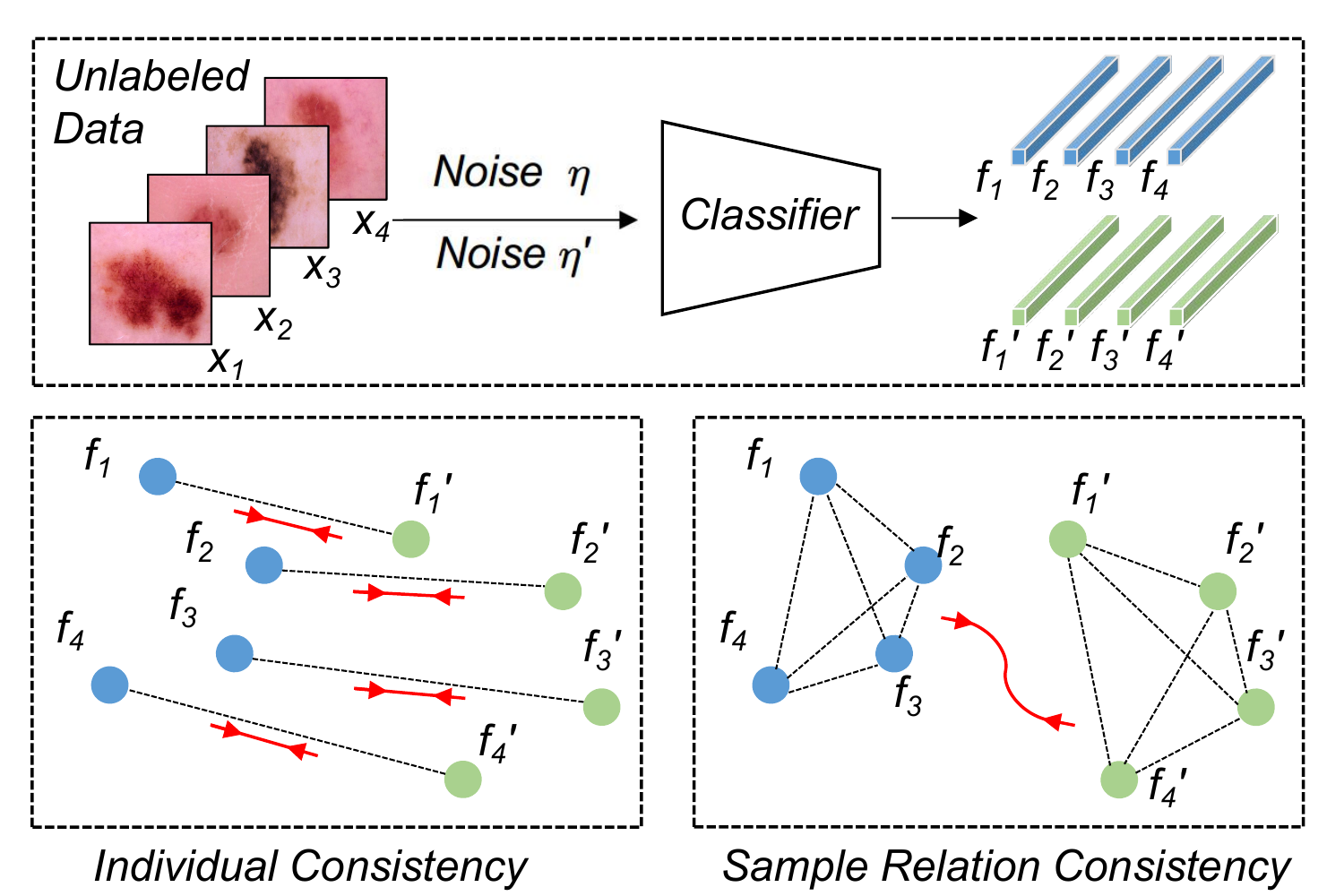}
\caption{\revisemajor{Conventional individual consistency and our proposed sample relation consistency (SRC) paradigms. The $f_i$ and $f_i'$ (blue/green points) denote the features extracted from the same input sample $x_i$ under different perturbations $\eta$ and $\eta'$, respectively. 
The consistency based methods emphasize the consistency (red arrows) between $f$ and $f'$ to explore the knowledge from unlabeled data. 
The individual consistency typically enforces the prediction consistency case by case, while the SRC further emphasizes the consistency of the intrinsic relation among different samples to extract additional semantic information from unlabeled data.}}
	\label{fig:PRC}
	\vspace{-3mm}
\end{figure}

Semi-supervised learning has shown potential for improving network performance when labeled data is scarce. Currently, most of the semi-supervised methods are following \textit{consistency enforcing} strategy~\cite{laine2017temporal,tarvainin2017mean}, which leverages the  unlabeled data by regularizing the network predictions to be consistent under input or network weight perturbations. 
Given an input sample, consistency enforcing methods create different perturbed samples of the same input (\eg, by adding Gaussian noise), and then encourage the model predictions on these samples to be similar~\cite{laine2017temporal}.  
In this stream of solutions,
previous consistency-based methods mainly focus on improving the quality of consistency targets (\ie~predictions of perturbed samples). For example, the Temporal Ensembling (TE) method~\cite{laine2017temporal} adopts an exponential moving average (EMA) of predictions in different epochs as the consistency targets. However, this method requires to maintain a huge prediction matrix during network training. 
To address this limitation, the Mean Teacher (MT)~\cite{tarvainin2017mean} framework constructs an ensembled teacher model for generating reliable consistency targets on the fly, which has been demonstrated with state-of-the-art performance on a variety of applications~\cite{tarvainin2017mean,cui2019semi,perone2018deep,yu2019uncertainty}.

One limitation of existing consistency based methods is the negligence of the relations among samples, which can benefit extraction of useful semantic information from unlabeled data. Recent studies have revealed that an entity,~\ie, an image, would become more informative when considering its intrinsic relation with other samples~\cite{liu2019knowledge, Battaglia2018relational}. 
For example, Liu~\etal~\cite{liu2019knowledge} \revisemajor{demonstrated} that the instance relation graph embedded extra semantic knowledge compared with individual instance representations in model compression. Meanwhile, such kind of relation information also widely exists in medical imaging. In clinical diagnosis, the experienced clinicians tend to make diagnosis according to the reference from previous analogous cases~\cite{avni2011xray}. 
Based on the above observations, the central tenet of our work is to explore such intrinsic relation between samples in an unsupervised manner to effectively exploit the semantic information from unlabeled data.

In this paper, we present a novel semi-supervised framework for medical image classification by utilizing the relation between different samples. 
Our framework is based on the state-of-the-art consistency-based strategy~\cite{tarvainin2017mean}, which enforces the prediction consistency of given samples under different perturbations. 
We further develop a self-ensembling teacher model to improve the quality of consistency targets. 
\revisemajor{To better exploit the valuable relation information of unlabeled data}, we propose a novel~\textit{Sample Relation Consistency} (SRC) paradigm in our semi-supervised learning framework, by explicitly enhancing the consistency of the intrinsic relation among different samples.
As shown in Fig.~\ref{fig:PRC}, different from previous methods which simply enforce the sample-level prediction consistency, our SRC encourages the consistency of the structured relation (similarity) among different samples, highlighting that samples with high similarity should also be highly related after adding perturbations. 
Specifically, given the input mini-batch of $n$ images, we model the intrinsic relation among different samples by calculating the $n \times n$ correlation matrix in the high-level semantic feature space. The SRC then minimizes the difference of such relation \revisemajor{matrices} under different perturbations, aiming to encourage the network to learn more robust and discriminative representation. 
The SRC regularization is independent from the label information, thus helping to extract extra semantic information from the unlabeled data for higher performance gains.
%
Our main contributions are summarized as follows:



\begin{enumerate}
	\item
	We present a new relation-driven semi-supervised framework for medical image classification, with general effectiveness on both single-label and multi-label classification tasks.
    \item
    We propose a novel \textit{sample relation consistency} paradigm to extract additional semantic information from the unlabeled data by explicitly enforcing the consistency of the relations among different samples under perturbations.
	\item
    We have conducted extensive experiments on two large-scale public benchmark medical image datasets, \ie~skin lesion diagnosis with ISIC 2018 challenge and thorax disease classification with ChestX-ray14.
    The results demonstrate consistent superior performance of our framework over state-of-the-art semi-supervised learning methods. 
    Code of our method is available at: \url{https://github.com/liuquande/SRC-MT}
\end{enumerate}


%
\section{Related work}
We first review the consistency-based semi-supervised methods which are closely relevant to our work, then recall the literature of semi-supervised learning in medical image analysis. 
%
Finally, we introduce the work for skin lesion classification and thorax disease diagnosis, on which we evaluate the effectiveness of our method.
\subsection{Consistency-based Semi-supervised Learning}
The consistency-based methods leverage unlabeled data by enforcing the prediction consistency under different perturbations. 
Most previous works focus on designing effective approaches to produce high-quality consistency targets (predictions of perturbed samples).
For example, the $\Pi$ model~\cite{laine2017temporal} directly utilizes the network outputs as the consistency targets. 
The Temporal Ensembling (TE) method~\cite{laine2017temporal} further adopts an exponential moving average (EMA) predictions for each unlabeled data as the consistency targets, which could effectively improve the quality of consistency targets due to the ensembling information from previous epochs. However, the TE requires to maintain a huge predictions matrix during training, making it heavy when learning from large datasets. 
To address this limitation, instead of maintaining the EMA predictions, the Mean Teacher (MT) framework constructs a teacher model with the EMA weights of the student model. The predictions of the teacher model could then serve as reliable consistency targets due to the effect of model ensembling. 
Based on these strategies, some recent works study more effective perturbation functions to improve the benefits from consistency regularization. For example, Xie~\etal~\cite{xie2019semi} \revisemajor{demonstrated} that utilizing better augmentation methods to create perturbed samples leads to greater improvements. \revisemajor{Miyato~\etal~\cite{miyato2019virtual} proposed the virtual adversarial perturbation to enhance the local smoothness of the label distribution given input samples. }
Different from these works above, we aim to improve existing consistency-based methods by exploring the intrinsic relation information of input data. 
\subsection{Semi-supervised Learning in Medical Image Analysis}
Semi-supervised learning has been studied in medical imaging community for a long period~\cite{cheplygina2018not,bai2017semi,zhang2017deep,jin2019incorporating}, for reducing the human effort on labelling data. 
The self-training approaches, which iteratively update the network parameters and the pseudo label for unlabeled data with an expectation-maximization procedure, have been utilized in a variety of medical image analysis tasks~\cite{ gu2017semi,singh2011identifying,bai2019self}.
Besides, co-training is also introduced for liver segmentation~\cite{xia2018cotraining} and breast cancer analysis~\cite{sun2016computerized} in a semi-supervised manner, where multiple classifiers are trained with independent sets of features and the classifiers rely on each other for estimating the confidence of their predictions. 
Some other approaches extract discriminative information from unlabeled data via feature space alignment~\cite{baur2017semi} and image reconstruction~\cite{chen2019discriminative}, achieving promising progress in sclerosis lesion segmentation and left atrium segmentation tasks.
%

More recent semi-supervised learning methods in \revisemajor{medical} image analysis domain can be grouped into three categories: 1) adversarial-learning-based approach~\cite{chartsias2018factorised, nie2018asdnet}, 2) graph-based approach~\cite{aviles2019graph}, and 3) consistency-based approach~\cite{li2020transformation}. 
For the first category, Dong~\etal~\cite{dong2018unsupervised} \revisemajor{introduced} adversarial-learning for semi-supervised lung segmentation, based on the assumption that unlabeled data has similar segmentation masks as labeled data. 
\revisemajor{Meanwhile, Diaz-Pint~\etal~\cite{andres2019retinal} introduced the generative adversarial network into glaucoma assessment, where both labeled and unlabeled data is used to train an image synthesizer for data augmentation.}
As a typical example for graph-based methods, \revisemajor{Aviles-Rivero~\etal~\cite{aviles2019graph,aviles2020when} constructed a graph model for thorax disease diagnosis from chest X-rays under extreme limited supervision, where the pseudo labels for unlabeled data are assigned via label propagation.}
Recently, the consistency-based methods have been successfully extended to medical imaging domain.  
For example, Li~\etal~\cite{li2018semi} \revisemajor{extended} the $\Pi$ model~\cite{laine2017temporal} for semi-supervised skin lesion segmentation with a transformation consistency strategy. Drawing spirit from Mean Teacher, other approaches~\cite{cui2019semi, perone2018deep,yu2019uncertainty} \revisemajor{enforced} the prediction consistency between the student model and a self-ensembling teacher model. 
Su~\etal~\cite{su2019local} \revisemajor{improved} the consistency based methods by constraining the feature space to learn more separable inter-class features and more compact intra-class features. 


\subsection{Skin Lesion Classification}

Accurate classification of skin lesions, especially melanoma recognition, is essential for assisting clinical diagnosis. 
Due to the inter-class similarity and intra-class variation of skin lesions, lots of efforts have been dedicated on addressing this challenging problem~\cite{ganster2001ganster,codella2015deep,yu2016automated,xue2019robust,zhang2019attention,shi2019active}. 
Early studies~\cite{ganster2001ganster} apply hand-crafted features,~\eg, shape, color and texture, to distinguish different lesions types, while depending heavily on the quality of extracted features. 
%
More recent works make use of the remarkable representation learning ability of convolutional neural networks (CNNs) to solve this problem. 
For example, Codella~\etal~\cite{codella2015deep} integrated CNNs, \revisemajor{sparse coding} and support vector machine for melanoma recognition. Yu~\etal~\cite{yu2016automated} \revisemajor{proposed} a very deep residual network to distinguish melanoma from non-melanoma lesions. 
%
%
Zhang~\etal~\cite{zhang2019attention} \revisemajor{proposed} an attention residual model that exploits self-attention ability of CNNs to focus on semantically meaningful regions of lesion.
Shi \etal~\cite{shi2019active} have recently proposed an active learning method for reducing annotation efforts, however, semi-supervised learning has not been explored within this scenario.

\subsection{Thorax Disease Diagnosis}
Establishing automatic diagnosis platform to understand chest radiography is of great significance in clinical practice.
To this end, nearly million-level chest x-rays datasets~\cite{wang2017chest8, irvin2019chexpert} have been developed recently to support relevant studies. 
%
%
Early studies on this task are based on handcrafted classification, which depend heavily on the quality of extracted feature. 
With the appearance of large scale dataset, more researchers adopt CNNs for solving this challenging task~\cite{wang2017chest8,irvin2019chexpert,yao2017learning,rajpurkar2017chexnet}.
For example, Wang~\etal~\cite{wang2017chest8} \revisemajor{demonstrated} the feasibility to detect and even localize the thorax disease from chest radiography using a multi-label learning framework.
%
%
After the releasing of ChestXray14~\cite{wang2017chest8} dataset, Rajpurkar~\etal~\cite{rajpurkar2017chexnet} \revisemajor{proposed} the state-of-the-art model,~\ie, CheXNet, to detect 14 common thorax diseases. 
However, the effectiveness and accuracy of these aforementioned approaches rely heavily on a large corpus of labelled data for network training.
To address this problem, Aviles-Rivero~\etal~\cite{aviles2019graph,aviles2020when} have recently proposed a graph-based semi-supervised framework for X-ray classification under extreme limited supervision. 

\begin{figure*}[t]
	\centering
	\includegraphics[width=1.0\textwidth]{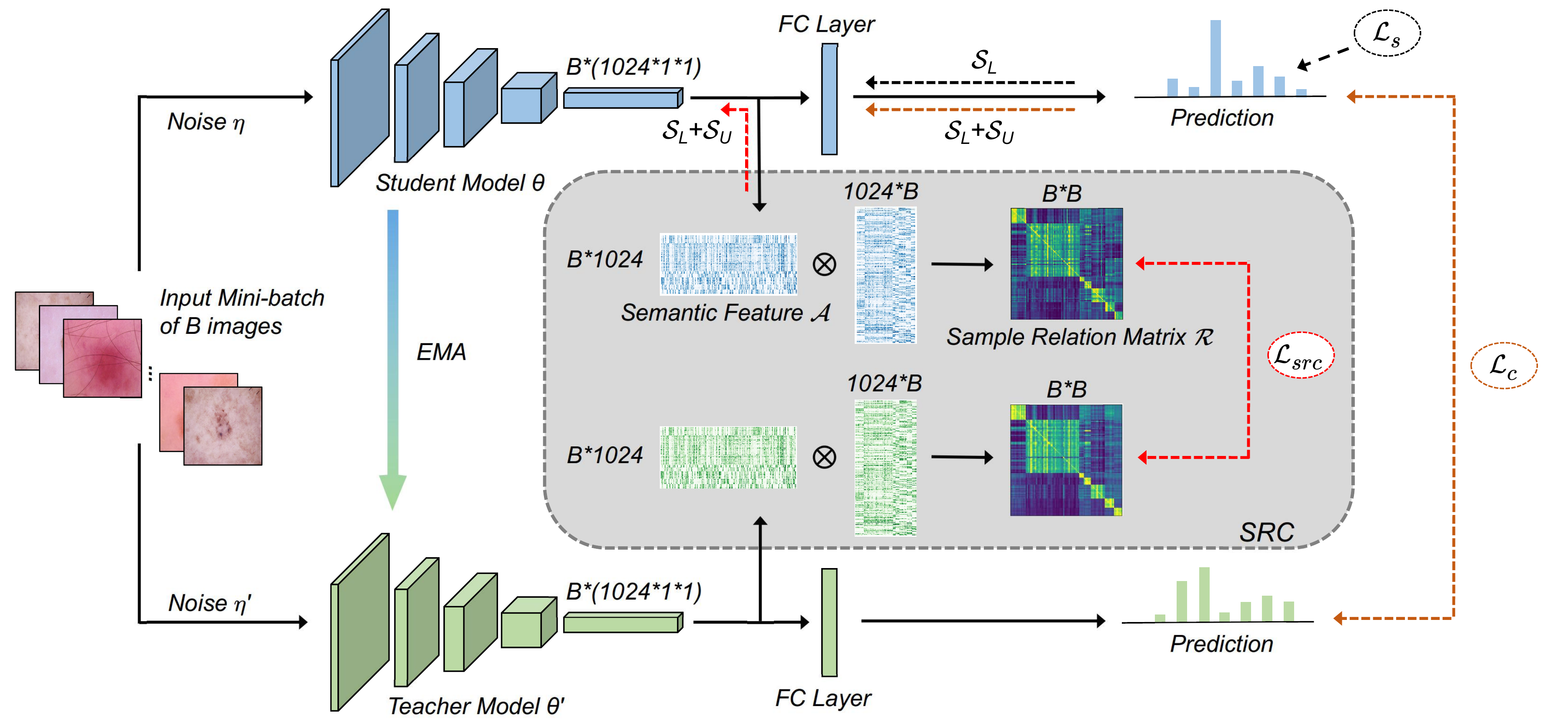}
	\caption{The overview of our relation-driven framework for semi-supervised medical image classification. The teacher weights $\theta'$ are updated as the exponential moving average (EMA) of student weights $\theta$. The objective function to optimize the student model includes the supervised loss ($\mathcal{L}_s$) on labeled set $\mathcal{S}_{L}$ and two unsupervised consistency loss ($\mathcal{L}_c$ and $\mathcal{L}_{src}$) on both unlabeled set $\mathcal{S}_U$ and labeled set $\mathcal{S}_L$. 
	The sample relation consistency (SRC) paradigm calculates the relation matrix from the semantic feature space to estimate the semantic relation between different samples. 
	The whole framework regularizes such relation matrix to be stable between student model and teacher model under different perturbations $\eta'$ and $\eta$ of the input by minimizing $\mathcal{L}_{src}$.}
	\label{fig:framework}
\end{figure*}
\section{Method}
The Fig.~\ref{fig:framework} depicts our proposed sample-relation-consistent mean-teacher framework (SRC-MT) for semi-supervised medical images classification. The sample relation consistency paradigm models the intrinsic relation among samples by calculating the relation matrix from the high-level semantic feature of each sample. Our framework regularizes this relation structure to be stable between teacher model and student model under different perturbations to extract richer semantic information from the unlabeled data. 
%
\subsection{The Backbone Semi-supervised Learning Framework}
%
%
We denote the labeled set as $\mathcal{S}_L = \{(x_i,y_i)\}_{i=1}^N$ and the unlabeled set as $\mathcal{S}_U = \{x_i\}_{i=N+1}^{N+M}$, where $x_i$ is the input 2D medical \revisemajor{image},~\eg, dermoscopy \revisemajor{image} or \revisemajor{chest X-ray}, $y_i$ is the one hot ground-truth label. 
\revisemajor{Our semi-supervised learning framework roots in the consistency regularization mechanism, by imposing various perturbations onto the same inputs and regularizing the consistency of model predictions to explore the ``dark knowledge"~\cite{laine2017temporal} from the unlabeled data.}
The total optimization objective of the whole framework can be formulated as following:
\begin{equation}
\underset{\theta}{\textrm{min}} \sum_{i=1}^N \mathcal{L}_s(f(x_i;\theta), y_i) + \lambda  \mathcal{L}_u (\{x_i\}_{i=1}^{N+M} ; f(\cdot), \theta, \eta, \theta', \eta')), 
\end{equation}
%
where $\mathcal{L}_s$ denotes the supervised loss (\eg, cross-entropy loss) for evaluating the network outputs on labeled inputs; 
$\mathcal{L}_u$ represents the unsupervised consistency loss to measure the consistency of the same inputs under different perturbations. 
\revisemajor{Here, $f(\cdot)$ denotes the classification network; $\theta$ and $\theta'$ are the parameters weights of the student model and teacher model respectively; while $\eta$ and $\eta'$ represent the different input perturbations (\eg, adding Gaussian noise) applied to the two models respectively.}
$\lambda$ is a ramp-up weighting factor that controls the trade-off between the supervised and unsupervised loss. 

Recent progress~\cite{tarvainin2017mean} on semi-supervised learning reveals that ensembling weights of student model at different training process helps build a more reliable teacher model to produce consistency targets. Inspired by this observation, in our framework, we employ a teacher model and update its weights $\theta'$ as the \textit{exponential moving average} (EMA) of the weights $\theta$ of student model and only optimize the student model in network training. 
Specifically, the teacher weights $\theta'_t$ at training iteration $t$ are updated as following:
\begin{equation}
\theta'_t = \alpha \theta'_{t-1} + ({1-\alpha}) \theta_t,
\end{equation}
where $\alpha$ is a smoothing coefficient hyper-parameter that controls the weights updating rate.

To encourage the consistency of teacher model and student model outputs, our framework preserves the conventional individual consistency mechanism~\cite{tarvainin2017mean, laine2017temporal}, which emphasizes the prediction consistency of each individual sample under different perturbations. This mechanism is expressed as following:
\begin{equation}
\mathcal{L}_c = \sum _{i=1} ^{N+M} \mathbb{E}_{\eta',\eta} \left\|f(x_i,\theta',\eta')-f(x_i,\theta,\eta)\right\|_2^2,
\end{equation}
where $x_i$ denotes each training sample.

\subsection{Sample Relation Consistency (SRC) Paradigm}
Recall from the introduction and recent study on graph neural network~\cite{Battaglia2018relational}, we can learn extra semantic information for an entity,~\eg, an radiography image, when considering the relation with other analogous entities. 
Based on this assumption, we propose to model such intrinsic relation to a consistency paradigm,~\ie, sample relation consistency, which regularizes the network to maintain the consistency of the semantic relation between samples under different perturbations, \revisemajor{and thereby encourage the network to explore additional semantic information from the input data to improve the network performance}. 
%

%
We model the structured relation among different samples with a case-level Gram Matrix~\cite{gatys2015gram}. 
Given an input mini-batch with $B$ samples, we denote the activation map of layer $l$ as $F^l \in \mathbb{R} ^{B\times C\times H \times W}$, where $H$ and $W$ are the spatial dimension of feature map, and $C$ is the channel number. 
We reshape the feature map $F^l$ into $A^l \in \mathbb{R} ^{B\times HWC}$, and then the Case-wise Gram Matrix $G^l \in \mathbb{R}^{B \times B}$ is computed as:
\begin{equation}
G^l = A^l \cdot (A^l)^T,
\end{equation}
where  $G_{ij}$ is the inner product between the vectorized activation map $A^l_{(i)}$ and $A^l_{(j)}$, whose intuitive meaning is the similarity between the activations of $i_{th}$ sample and $j_{th}$ sample within the input mini-batch. 
The final sample relation matrix $R^l$ is obtained by conducting the L2 normalization for each row $G_i^l$ of $G^l$, which is expressed as:
\begin{equation}
\label{eq:relationmatrix}
R^l = [\frac{G^l_1}{\left\|G^l_1\right\|_2},...,\frac{G^l_B}{\left\|G^l_B\right\|}_2]^T.
\end{equation}
%
%

The SRC requires the relation matrix $R^l$ to be stable under different perturbations to preserve the semantic relation between samples.
We then define the proposed SRC-loss as:
\begin{equation}
\label{eq:prcloss}
\mathcal{L}_{src} = \underset{\mathcal{X} \in \{\mathcal{S}_U\cup\mathcal{S}_L\}}{\sum}  \frac{1}{B} \left\|R^l(\mathcal{X}; \theta, \eta) - R^l(\mathcal{X}; \theta', \eta')\right\|_2^2,
\end{equation}
where $\mathcal{X}$ is the sampled mini-batch from training set $\{\mathcal{S}_U\cup\mathcal{S}_L\}$, $R^l(\mathcal{X}; \theta, \eta)$ and $R^l(\mathcal{X}; \theta', \eta')$ are the sample relation matrices computed on  $\mathcal{X}$ under different wights and perturbations pair $(\theta, \eta)$ and $(\theta', \eta')$, respectively. 
By minimizing $\mathcal{L}_{src}$ during the training process, 
the network would be enhanced to capture more robust and discriminative representation that benefits for preserving the intrinsic relation between samples under different perturbations, thus helping to extract additional semantic information from unlabeled data.

\begin{figure}[t]
	\centering
	\includegraphics[width=0.48\textwidth]{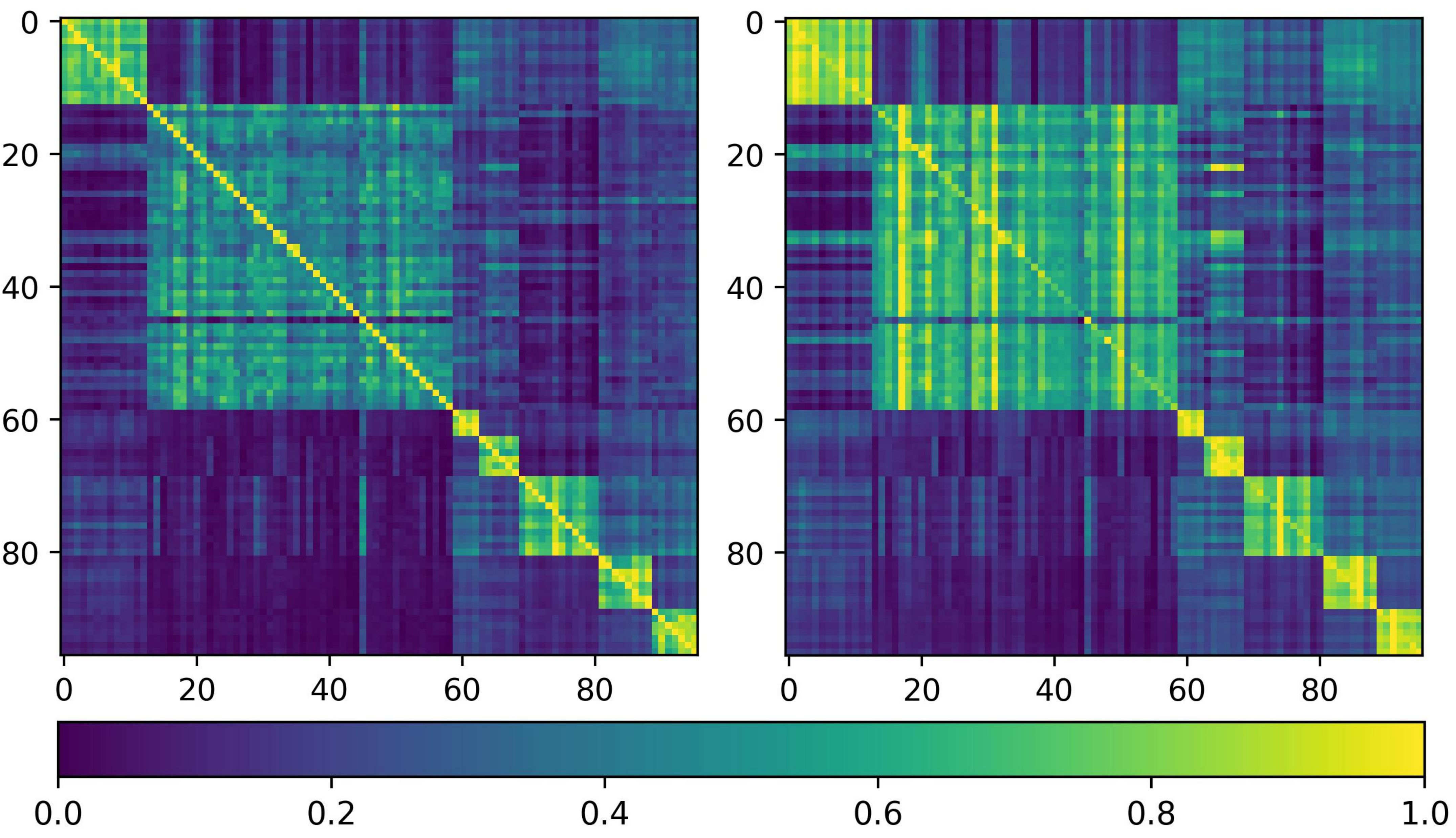}
	\caption{Sample relation matrix $R^l$ calculated with the same mini-batch (batch size = 96) from the ISIC 2018 skin lesion classification dataset. The two sub-figures show $R^l$ calculated with the features before (left) and after (right) the last pooling layer of a fully-supervised classification network. For both two sug-figures, the $i$-th row in the figure represents the similarity of the $i$-th sample with all samples within this mini-batch. Note that the batch of samples have been grouped by their ground truth label class along each axis before visualization.}
	\label{fig:activationsplace}
	\vspace{-2mm}
\end{figure}
We also study which layer to choose for estimating the semantic relation among different samples. Generally, the activations from deeper layers are preferable in our method, as they contain more high-level semantic information than the activations from middle layers. 
We therefore only compare the sample relation matrix $R^l$ (see Eq.~\eqref{eq:relationmatrix}) calculated using the activations before and after the last global average pooling layer \revisemajor{(\ie, the pooling layer before the fully connected layer in the employed DenseNet~\cite{huang2017dense})}.
As shown in Fig.~\ref{fig:activationsplace}, the left and right figures denote the $R^l$ from the same sampled mini-batch, but calculated with the features before and after the last pooling layer, respectively. 
For each of the two sub-figures, the $i$-th row represents the similarity between the $i$-th sample and all samples in this mini-batch. 
Note that the input batch of samples have been grouped by their ground truth class to better observe the relation among samples.
The block-wise patterns in the sub-figures indicate that the samples of the same class have high similarity in their semantic feature space. 
In addition, we observe that the right sub-figure has higher response in the diagonal boxes than the left one, indicating that the semantic relation among different samples are better presented in the feature space after the last pooling layer. 
The reason maybe that the last pooling layer helps reduce the spatial difference of the semantic features. Suppose two images have the same lesion type while the lesions appear on different positions in the images, then the similarity calculated with activations before last pooling layer would be relatively small due to the spatial discrepancy. 
Therefore, we choose the activations after the last pooling layer to calculate $R^l$ in our framework. 

\begin{table*}
    \renewcommand\arraystretch{1.2}
	\centering
    \caption{Comparison with state-of-the-art semi-supervised learning methods on ISIC 2018 dataset.}
    \label{tab:tab1}
    \begin{tabular}{c|cc|ccccc}
        \hline
        \multirow{2}{*}{Method} & \multicolumn{2}{c|}{ Percentage} &\multicolumn{5}{c}{Metrics} \\
        \cline{2-8}
                    &Labelled  &Unlabelled    &AUC    &Sensitivity    &Specificity   &Accuracy    &F1    \\
        \hline
        Upper Bound &100\%   &0           &95.43  &75.20          &94.94         &95.10       &70.13 \\
        Baseline    &20\%    &0           &90.15  &65.50          &91.83         &92.17       &52.03 \\
        \hline
        Self-training~\cite{bai2017semi} 
                    &20\%   &80\%       &90.58  &67.63          &\textbf{93.31} &92.37 & 54.51\\
        SS-DCGAN~\cite{andres2019retinal} &20\% &80\% &91.28 &67.72 & 92.56 &92.27 &54.10\\
        
        TCSE~\cite{li2018semi} 
                    &20\%   &80\%       &92.24  &68.17          &92.51          &92.35          &58.44 \\
                    
        TE~\cite{laine2017temporal} &20\% &80\% &92.70 &69.81 &92.55 &92.26 &59.33\\
        \hline
        MT~\cite{tarvainin2017mean} 
                    &20\% &80\%       &92.96  &69.75          &92.20          &92.48          &59.10\\
      \textbf{SRC-MT (ours)} 
                    &20\%   &80\%       &\textbf{93.58} &\textbf{71.47} &92.72  &\textbf{92.54} &\textbf{60.68}\\

        \hline
    \end{tabular}
    \vspace{-3mm}
\end{table*}

\subsection{Overall Loss Function and Technical Details}
\revisemajor{The total objective functions to train our relation-driven semi-supervised framework is as follows:
\begin{equation}
\label{eq:overallloss}
\mathcal{L} = \mathcal{L}_s + \lambda \mathcal{L}_u, ~~\text{with}~~ \mathcal{L}_u = \mathcal{L}_c + \beta \mathcal{L}_{src},
\end{equation}
where $\mathcal{L}_s$ is the supervised objective; $\mathcal{L}_u$ is the unsupervised objective composed of the conventional consistency loss $\mathcal{L}_{c}$ and the sample relation consistency loss $\mathcal{L}_{src}$; $\beta$ is a hyper-parameter to balance $\mathcal{L}_c$ and $\mathcal{L}_{src}$, which is generally set as 1 and we also study the effect of this hyper-parameter in ablation experiments; $\lambda$ is the trade-off weight between the supervised and unsupervised loss.}

We \revisemajor{used} the weighted cross-entropy loss as $\mathcal{L}_s$ to mitigate the class imbalance problem for the two tasks in our experiments. \revisemajor{Two types of perturbations were added during training, including: 1) Random rotation, translation, horizontal flips to input samples:
We conducted random transformation for each sample separately in the given mini-batch. The random rotation degree was in range of -10 to 10. The pixel number for horizontal and vertical translation ranged from -2\% to 2\% of the image width. We also conducted random horizontal and vertical flipping for the inputs with a 50\% probability; 2) Dropout layer in the network: We added dropout layer before the last pooling layer in the employed DenseNet, with dropout rate as 0.2. The magnitude of these perturbations were chosen based on our empirical observations in preliminary experiments.}
We \revisemajor{turned} on the dropout during training while \revisemajor{turned} it off during validation and testing phase. 
We set the EMA decay rate $\alpha$ as 0.99, following~\cite{tarvainin2017mean}. \revisemajor{We applied a Gaussian warming up function $\lambda (t) = 1 * e ^ {(-5(1-t/T)^2)}$ to control the value of trade-off weight $\lambda$. The function value would gradually ramp-up from 0 to 1 in the first T training epochs, and then we fixed the value of $\lambda$ as 1 for the subsequent training.
Such design could guarantee that the training loss would not be dominated by the unsupervised loss at the beginning of network training when the consistency targets for unlabeled data are unreliable.}
\section{Experiments}
We have evaluated our proposed semi-supervised learning approach on both skin lesion classification (single-label) from dermoscopy images and thorax diseases diagnosis (multi-label) from chest x-ray images, with extensive ablation analysis and comparison with state-of-the-art methods.

%
\subsection{Datasets and Experimental Setup}
%
%

\emph{\textbf{ISIC 2018 skin lesion analysis dataset}}: We \revisemajor{performed} single-label skin lesion classification on the dataset from ISIC 2018: Skin Lesion Analysis Towards Melanoma Detection~\cite{codella2018skin, tschandl2018the}. 
The training set consists of 10015 skin lesion dermoscopy images, labeled by 7 types of common skin lesions. 
We resized all images into the size of 224 $\times$ 224. To employ the pre-trained model, each image was normalized with statistic collected from ImageNet dataset~\cite{russa2017imagenet} before feeding into the network. 
Since the ground truth of official validation and testing set was not released, we randomly divided the entire training set to 70\% for training, 10\% for validation and 20\% for testing. We employed DenseNet121~\cite{huang2017dense} pre-trained on ImageNet as our network backbone.

\emph{\textbf{ChestX-ray14 dataset}}: We \revisemajor{performed} multi-label thorax disease diagnosis using the ChestX-ray14 dataset~\cite{wang2017chest8}. It contains in total 112120 frontal-view chest x-rays of 30805 unique patients, in which each radiography is labeled with one or multiple classes of 14 common thorax diseases. We resized the original images from size 1024 $\times$ 1024 into 384 $\times$ 384 for training a deep neural network. The images were normalized in the same way as task 1. 
To ensure the fair comparison with previous methods~\cite{rajpurkar2017chexnet, aviles2019graph}, we adopted the official data split of ChestX-ray14, which split the entire dataset into 70\% for training, 10\% for validation and 20\% for testing. Since this dataset is much larger than the dataset in task-1, we adopted a deeper network, \ie, DenseNet169~\cite{huang2017dense} pre-trained on ImageNet, as our network backbone.

\emph{\textbf{Evaluation metric}}: Following the literature of these two applications, we adopt four metrics in the evaluation of ISIC 2018 dataset, including AUC, Accuracy, Sensitivity and Specificity. For the evaluation on ChestX-ray14 dataset, we refer to the previous work~\cite{aviles2019graph} and adopt AUC metric for evaluation.

\emph{\textbf{Implementation details}}: Our framework was implemented in Python with PyTorch library.
We used 3 NVIDIA TitanXp GPUs in parallel for training. \revisemajor{The network was trained using Adam optimizer with default parameters setting~\cite{kingma2015adam}.} 
The batch size was set to 48, containing 12 annotated images and 36 unannotated images in each mini-batch.
\revisemajor{We totally trained 60 epochs for skin lesion classification and 20 epochs for ChestX-ray diagnosis, with ramp-up epoch $T$ set as 30 and 10 respectively}. 
The learning rate was initialized as $1e^{-4}$ and decayed with a power of 0.9 after each epoch.

\subsection{Comparison on Skin Lesion Classification Dataset}
We implemented current state-of-the-art semi-supervised classification methods on the skin lesion classification dataset for comparison, including self-training based method~\cite{bai2017semi}, GAN based method~\cite{andres2019retinal}, $\Pi$ model based method~\cite{li2018semi} and Temporal Ensembling (TE)~\cite{laine2017temporal}. We also conducted comparison with the original mean teacher (MT) framework to evaluate the effect of our proposed SRC paradigm. \revisemajor{In self-training based method, we employed the model from previous epoch to generate soft probability and selected probability vectors with zero norm higher than 0.9 to produce high-confidence labels for unlabeled data.
The latent variable to generate fake images in GAN based method was a 256-dimensional vector sampled from Gaussian distribution. The generator was composed of six deconvolutional block (Deconv-BN-ReLU), followed by a deconvolutional layer of 3 output channel with Tanh activation function. We adopted mean squared error as the GAN loss function.
All consistency-based methods applied the same perturbations to the input samples, as elaborated in the technical details of our method.} In addition, all the comparison methods were implemented with the same network backbone for fair comparison. 

Table~\ref{tab:tab1} presents the performance of these approaches under 20\% (1400) labeled data setting. 
The fully supervised model trained with 100\% (7000) labeled data serves as the upper-bound performance and fully supervised model trained with 20\% (1400) labeled data serves as the baseline performance. 
As we can see, the self-training based method obtains higher Specificity than the other approaches, while the \revisemajor{improvements} over baseline model on AUC and Sensitivity \revisemajor{are} limited. Such result indicates that the self-training methods mainly helps to improve the classification for negative samples while the performance improvements on positive samples \revisemajor{are} relatively poor. 
Compared with self-training, the SS-DCGAN improves the AUC by 0.70\%, showing that the generated samples of GAN based methods indeed help to improve the network training in semi-supervised learning. %
The TCSE and TE achieve large improvements over SS-DCGAN, demonstrating the effectiveness of consistency regularization mechanism for exploiting the unlabeled data. 
Meanwhile, the TE performs slightly better than TCSE, since it ensembles the predictions in different epochs to generate more reliable consistency targets. 
In addition, the MT obtains a higher AUC of 92.96\% than all other methods, which is consistent with the findings in~\cite{tarvainin2017mean}, demonstrating the superiority of MT in semi-supervised learning. Notably, by using SRC to enforce the consistency of the semantic relation among different samples, our approach achieves consistent improvements on all metrics over MT, with considerable increases on AUC, Sensitivity and F1 score, highlighting that the proposed SRC paradigm indeed helps to exploit the unlabeled data more effectively. 
%
%
%
We also reference the state-of-the-art method~\cite{gessert2019skin} on skin lesion classification to validate our backbone implementation. They report performance of 96.7\% AUC for 5-fold cross validation on the training set, using Densenet121 as network backbone. Compared with their results, our upper-bound implementation with 95.43\% AUC can be regarded as reasonable.

\begin{table*}[h]
    \renewcommand\arraystretch{1.2}
	\centering
    \caption{Quantitative evaluation of our method on ChestXray14 dataset with AUC metric.}\label{tab4}
\begin{tabular}{l|p{2.2cm}<{\centering}|p{2.2cm}<{\centering}|p{2.2cm}<{\centering}|p{2.2cm}<{\centering}}
        \hline
        Method             &Fully-Supervised &Baseline   &MT~\cite{tarvainin2017mean} &SRC-MT\\
        \hline
        Labeled            &100\%    &20\%   &20\%   &20\%\\
        Unlabeled          &0       &0      &80\%   &80\%\\
        \hline
        Atelectasis        &77.32	&74.06	&75.12	&\textbf{75.38}\\
        Cardiomegaly       &88.85	&87.39	&87.37	&\textbf{87.70}\\
        Effusion           &82.11	&80.49	&80.81	&\textbf{81.58}\\
        Infiltration       &70.95	&69.78	&\textbf{70.67}	&70.40\\
        Mass               &82.92	&76.93	&77.72	&\textbf{78.03}\\
        Nodule             &77.00	&71.90	&73.27	&\textbf{73.64}\\
        Pneumonia          &71.28	&65.00	&69.17	&\textbf{69.27}\\
        Pneumothorax       &86.87	&82.58	&85.63	&\textbf{86.12}\\
        Consolidation      &74.88	&71.27	&72.51	&\textbf{73.11}\\
        Edema              &84.74	&80.84	&82.72	&\textbf{82.94}\\
        Emphysema          &93.35	&87.54	&88.16	&\textbf{88.98}\\
        Fibrosis           &84.46	&78.22	&78.24	&\textbf{79.22}\\
        Pleural Thickening &77.34	&74.50	&74.43	&\textbf{75.63}\\
        Hernia             &92.51	&84.91	&\textbf{87.74}	&87.27\\
        \hline
        Average AUC        &81.75   &77.52  &78.83  &\textbf{79.23}\\
        \hline
    \end{tabular}
\end{table*}

\begin{table}
   \renewcommand\arraystretch{1.2}
	\centering
    \caption{Comparison with state-of-the-art semi-supervised learning method on Chest X ray 14 dataset.}\label{tab:tab2}
    \begin{tabular}{l|c|c|c|c|c|c}
        \hline
        Labeled Percentage            &2\%    &5\%   &10\%   &15\% &20\% &100\%\\
        \hline
        GraphX$^{NET}$~\cite{aviles2019graph}   &53	&58	&63	&68 &78 &N/A\\
        SRC-MT(Ours)      &66.95	&72.29	&75.28	&77.76 &79.23 &81.75\\
        \hline
    \end{tabular}
    \vspace{-3mm}
\end{table}

\subsection{Comparison on Thorax Disease Diagnosis Dataset}
To evaluate our method on multi-label classification task, we compare our method with the graph-based model GraphX$^{NET}$~\cite{aviles2019graph}, which achieves the state-of-the-art semi-supervised performance on thorax disease diagnosis. 
The GraphX$^{NET}$ constructs a graph model with all training samples and then performs semi-supervised learning by correcting the initially mislabelled samples through finding smooth solutions to the created global embedding. 
We directly reference their reported performance for comparison as they also adopt the official data split of ChestX-Ray14~\cite{wang2017chest8}. 

Table~\ref{tab:tab2} shows the AUC of the GraphX$^{NET}$ and our method under different labelled data percentage. 
It is observed that the GraphX$^{NET}$ works pretty well under 20\% labeled data percentage, with 78\% AUC achieved on the official testing set, demonstrating the effectiveness of graph model for semi-supervised multi-label classification task. 
However, this method is very sensitive to the change of labeled data percentage. The AUC of GraphX$^{NET}$ catastrophically drops by 10\% when the labeled data percentage decreases from 20\% to 15\%, and even falls to 53\% under 2\% labeled data percentage. 
A possible reason is that the graph model relies heavily on a widespread labeled samples to ensure the correctness of the label propagation among labeled and unlabeled data.
Remarkably, our approach overwhelmingly outperforms the GraphX$^{NET}$ under all labeled data percentage. 
Compared with GraphX$^{NET}$, our method \revisemajor{presents} smaller performance decay when the labeled data percentage decreases, reflecting that our method is more robust to the fluctuations of labeled data percentage.
In addition, our method trained with only 5\% labeled data even outperforms the GraphX$^{NET}$ trained with 15\% labeled data, demonstrating that our method performs better than graph model under extreme limited supervision. 
\revisemajor{
We also validate our backbone implementation. To the best of our knowledge, the current state-of-the-art performance on ChestX-ray14 dataset is achieved by Yan~\etal~\cite{yan2018weakly}.  Their method obtains an average AUC of 83.02\% on the official data split, but relying on complicated network architecture design and testing model ensembling. They also report a 81.80\% average AUC using DenseNet as network backbone. Compared with their results, our implemented fully-supervised DenseNet with 81.75\% AUC on the official split can be regarded as valid.}

We also compare our method with the original MT framework to \revisemajor{analyze} the effect of SRC paradigm on the multi-label classification task. 
As shown in Table~\ref{tab4}, the MT framework trained with 20\% labeled data achieves average AUC of 78.83\%, with AUC improvements on 13 out of 14 types of thorax diseases (except Cardiomegaly) compared with the fully-supervised baseline model, demonstrating the effectiveness of consistency-based method on semi-supervised multi-label classification task. 
We notice that the performance gain of MT over baseline model on AUC (1.31\%) is not as significant as the performance gain in skin lesion classification (2.81\%), which could be due to the annotation noise in ChestXray14 dataset~\cite{wang2017chest8}. These noise in the training set would influence the discrimination ability of the trained model, which further decreases the quality of the consistency targets for unlabeled data and thus reduces the benefits from consistency regularization mechanism. 
%
Notably, by leveraging the SRC to explore the semantic relation information, our method further improves the average AUC to 79.23\%, with increase on 12 out of 14 diseases compared with MT framework, which could demonstrate the effectiveness of the proposed SRC paradigm on multi-label classification task.  
\begin{table*}[h]
    \renewcommand\arraystretch{1.2}
	\centering
    \caption{Quantitative evaluation of our method on ISIC 2018 dataset under different percetage of labeled data.}
    \label{tab:tab3}
    \begin{tabular}{c|cc|ccccc}
        \hline
        \multirow{2}{*}{Method} & \multicolumn{2}{c|}{Percentage} &\multicolumn{5}{c}{Metrics} \\
        \cline{2-8}
                    &Labelled  &Unlabelled    &AUC    &Sensitivity    &Specificity   &Accuracy    &F1    \\
        \hline
        Upper Bound &100\%   &0          &95.43  &75.20          &94.94         &95.10       &70.13 \\
        \hline
        Baseline    &5\%   &0 &84.24  &59.69 & 87.28&84.73 &38.57\\
        SRC-MT (ours) 
                    &5\%   &95\%       &87.61  &62.04 &89.36&88.77 &46.26\\
        \hline
        Baseline    &10\%   &0          &87.04  &64.22          &89.88         &87.45       &44.43 \\
        SRC-MT (ours) 
                    &10\%   &90\%       &90.31 &66.29 &90.47  &89.30 &50.02\\
        \hline
        Baseline    &20\%   &0          &90.15  &65.50          &91.83         &92.17       &52.03 \\
        SRC-MT (ours) 
                    &20\%   &80\%        &93.58 &71.47&92.72&92.54&60.68\\
        \hline
        Baseline    &30\%   &0 &91.80 &71.63 &92.78 &92.55  &57.83   \\
    SRC-MT (ours) 
                    &30\%   &70\%  &94.27 &74.59 &92.85 &93.11 &63.54  \\
        \hline
    \end{tabular}
\end{table*}

\begin{table*}[h]
    \renewcommand\arraystretch{1.2}
	\centering
    \caption{Quantitative evaluation of our method on ISIC 2018 dataset with different loss weight $\beta$.}
    \label{tab:tab5}
    \begin{tabular}{c|c|ccccc}
        \hline
        \multirow{2}{*}{Method} & \multicolumn{1}{c|}{hyper-parameter} &\multicolumn{5}{c}{Metrics} \\
        \cline{3-7}
                    & weight $\beta$    &AUC    &Sensitivity    &Specificity   &Accuracy    &F1    \\
        \hline
        MT
                    &0 &92.96 &69.75 &92.20 &92.48 &59.10\\
        \hline
        SRC-MT (ours) 
                    &0.01      &93.08 &70.64 &92.49 &92.57 &59.68\\
        SRC-MT (ours) 
                    &0.1      &93.47 &71.01 &92.20  &92.63 &59.88\\
        SRC-MT (ours) 
                    &0.5      &93.57 &70.45 &92.41  &92.90 &62.41\\
        SRC-MT (ours) 
                    &1.0       &93.58 &71.47&92.72&92.54&60.68\\
        SRC-MT (ours) 
                    &5.0     &93.39 &69.99 &92.25  &92.95 &60.15\\
        SRC-MT (ours) 
                    &10.0     &92.87 &68.66 & 92.51 &92.04 &58.50\\
        \hline
    \end{tabular}
\end{table*}

\subsection{Analytical Ablation Studies}
We provide ablation analysis on skin lesion classification dataset to further investigate the learning behavior of the proposed method.
\subsubsection{\textbf{Different percentage of labeled data}}
We study the impact of different percentage of labeled data in our semi-supervised learning method and show the results in Table~\ref{tab:tab3}. 
It is observed that our method achieves consistent improvements over the supervised-only baseline model under 5\%, 10\%, 20\% and 30\% labeled data setting.
In addition, using only 30\% labeled data for network training, our method achieves AUC of 94.27\% and Sensitivity of 74.59\% , which are close to the performance of upper-bound model trained on 100\% labeled data. These results demonstrate the effectiveness of our approach in exploiting the unlabeled data for performance gains. 
%
Moreover, we observe that the model trained with 5\% labeled data presents comparable performance gains over the supervised-only baseline as 20\% labeled data setting, even though we have utilized more unlabeled data for training in 5\% labeled data setting. One possible reason is that for the class imbalanced problem, \eg, skin lesion classification, using very limited labeled data (5\%, \ie, 350 images) for network training cannot guarantee reliable consistency targets for the unlabeled rare lesion types, which would reduce the benefits from consistency regularization mechanism.
%
%
\subsubsection{\textbf{The impact of different loss weight $\beta$}}
We study the effect of different hyper-parameter setting for $\beta$ in Eq.~\eqref{eq:overallloss}.
\revisemajor{Specifically, we adopt different values of $\beta$ in range from 0 to 10, and report the network performance under 20\% labeled data setting in Table~\ref{tab:tab5}. 
Note that the setting of $\beta = 0$ corresponds to the original MT framework. 
It is observed that our method generally improves the classification performance over MT framework with $\beta$ in range from 0.1 to 5, and the performance is not very sensitive to the value of $\beta$. When decreasing $\beta$ to 0.01, the improvement over MT is marginal, since the training loss would be dominated by the classification loss and the conventional consistency loss. 
Meanwhile, we notice a limitation from Table~\ref{tab:tab5} that the value of $\beta$ cannot be set too high (\eg, 10). Otherwise, the SRC paradigm might be a too strict constraint with negative effects. 
We thereby set $\beta$ as 1 in our experiments.}
%

\subsubsection{\textbf {Evolution of sample relation matrix}}
\begin{figure}[t]
	\centering
	\includegraphics[width=0.49\textwidth]{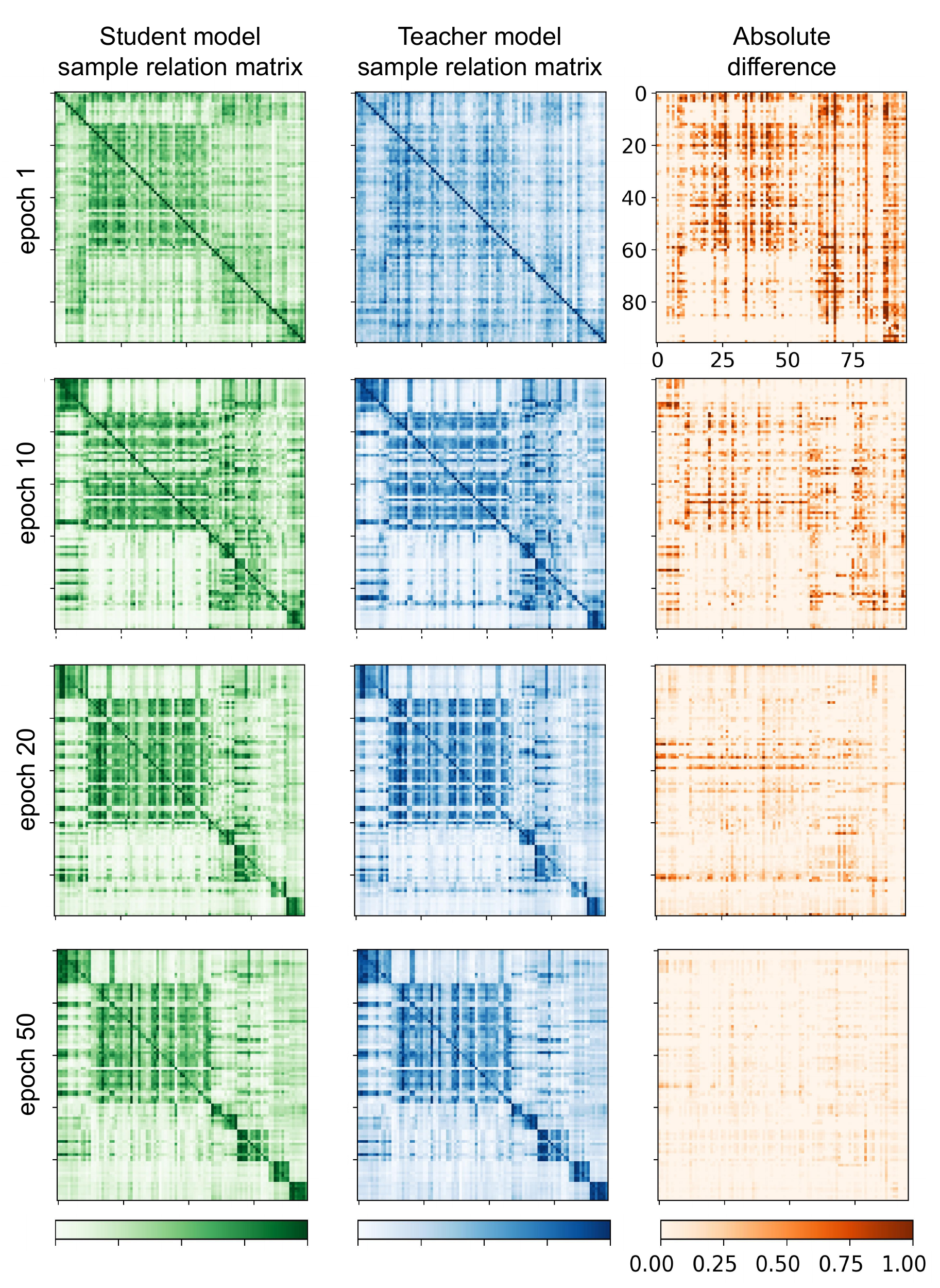}
	\vspace{-3mm}
	\caption{Visualization of the evolution of the sample relation \revisemajor{matrices} from student model (left column) and teacher model (middle column) as training going on. We also show the difference of these two relation \revisemajor{matrices} in the right column (with values amplified by three-times). 
	The first to fourth rows correspond to the results from 1, 10, 20, and 50 epochs during network training under 20\% labeled data setting.}
	\label{fig:distance}
	\vspace{-5mm}
\end{figure}

To better understand the learning behavior of SRC paradigm in network training, in Fig.~\ref{fig:distance}, we visualize the sample relation \revisemajor{matrices} $R^l$ calculated from the student model and teacher model in different training epochs.
To clearly show the alignment of these two \revisemajor{matrices}, we also compute their absolute distance matrix, see the right red column. 
From Fig.~\ref{fig:distance}, it is observed that at the beginning of network training, the intrinsic relation structure among different samples is not well presented (Note that samples in the mini-batch are grouped by their ground-truth label class), and the calculated relation \revisemajor{matrices} show large difference due to the input perturbations. 
As the training goes on, the model gradually produces meaningful relation \revisemajor{matrices}, with higher response between samples of the same lesion types. Meanwhile, the absolute difference between the two relation \revisemajor{matrices} gradually reduces as the model converges, indicating that the model gradually learns robust representations to preserve the semantic relation among samples under perturbations. 

\begin{figure}[t]
	\centering
	\revisemajor{
	\includegraphics[width=0.49\textwidth]{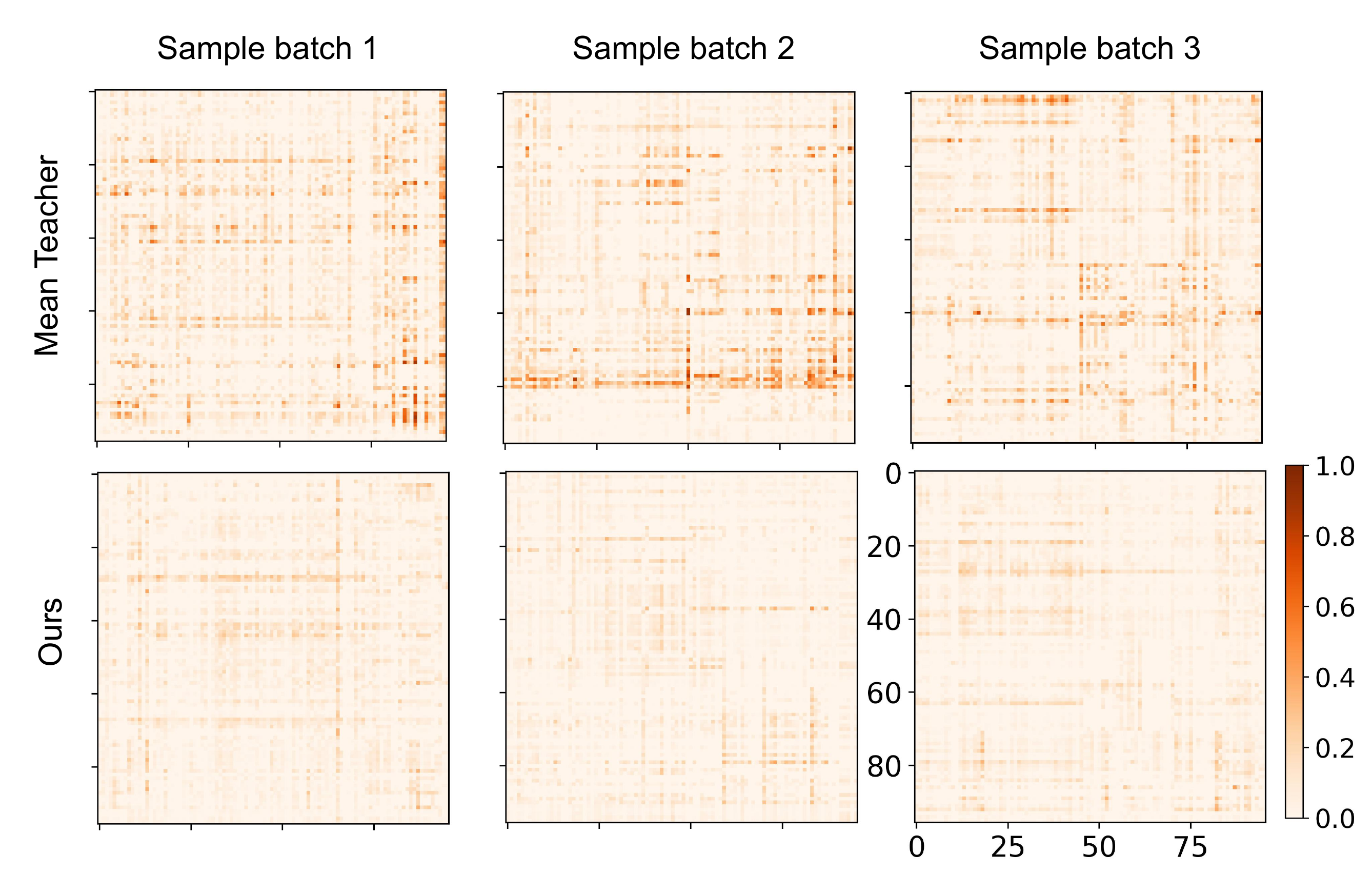}
	\vspace{-3mm}
	\caption{Visualization of the absolute distance (with values amplified by three-times) of the relation matrices $R^l$ calculated from the student model under two different perturbations, at 50th training epoch with 20\% labeled data setting.
	The first row and second row are the absolute distance matrices of mean teacher and our approach  respectively. Each column corresponds to the results from a certain sampled mini-batch.}
	\label{fig:absolute_distance}
	}
	\vspace{-3mm}
\end{figure}

\revisemajor{We also visualize the absolute distance of the mean teacher framework and our method at 50th training epoch when model converges to further analyze the impact of the SRC paradigm. As shown in Fig~\ref{fig:absolute_distance}, the absolute distance matrices in mean teacher framework present relatively high responses, reflecting that the sample relationships would be disturbed when the perturbations are introduced, even though the model has learned meaningful features at convergence. Notably, with the regularization from our SRC paradigm, the distance matrices turn cleaner, highlighting that our approach successfully captures more robust and discriminative representations that benefits for stabilizing the relation structure under perturbations. Such observation could also explain the performance improvements of our method.}

\begin{figure}[t]
	\centering
	\revisemajor{
	\includegraphics[width=0.47\textwidth]{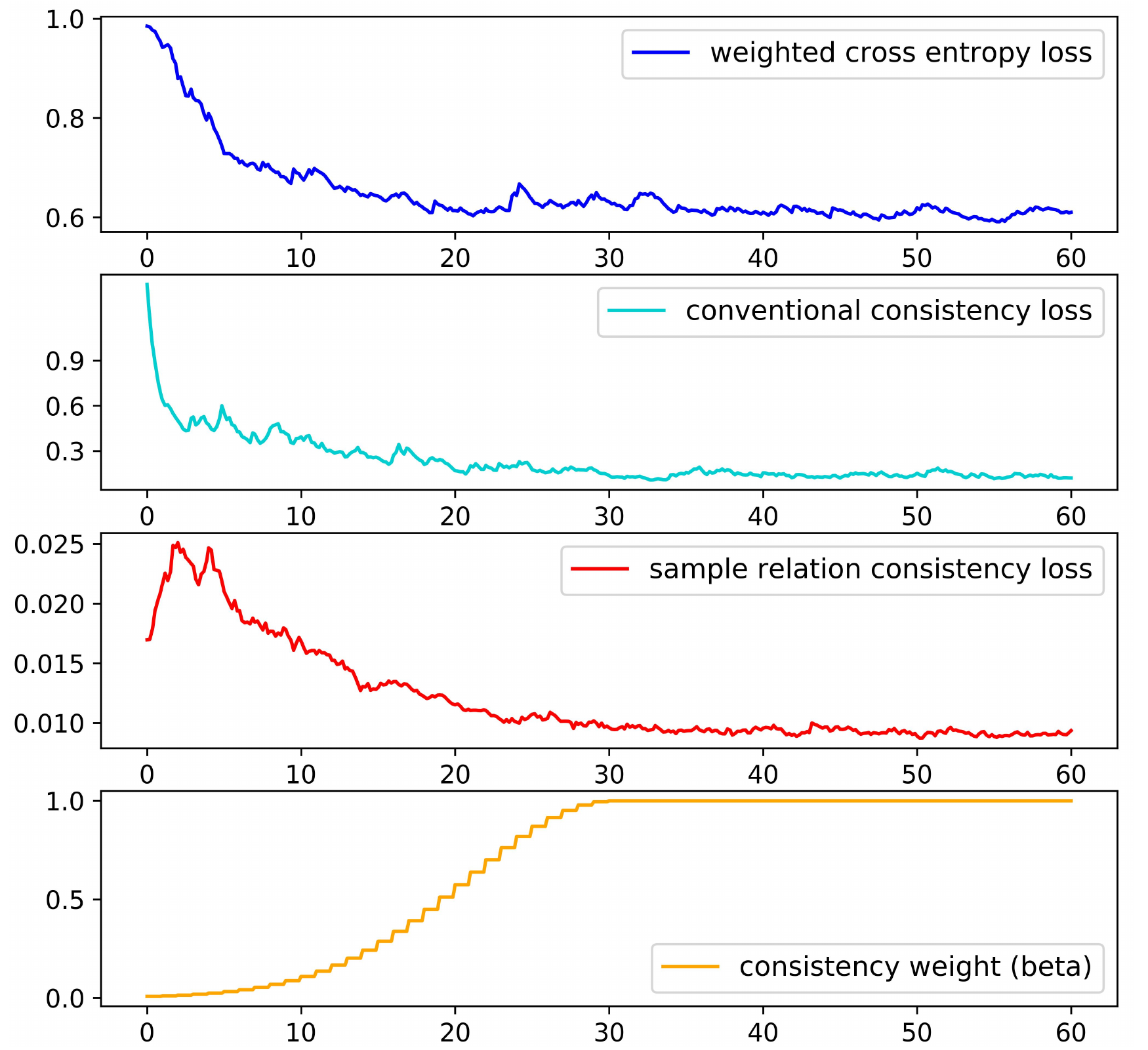}
   \caption{Learning curves of different loss functions and consistency weight $\beta$.}
	\label{fig:learning_curve}
}
	\vspace{-3mm}
\end{figure}

\revisemajor{
\subsubsection{\textbf{Analysis of training behavior}}
To analyze the training behavior of our model, we further show the training curves of the classification loss, conventional consistency loss, SRC loss and the consistency weight $\beta$. As shown in Fig.~\ref{fig:learning_curve}, the consistency weight $\beta$ gradually ramps up as the training goes on, which guarantees that the training would not be misled by the unreliable unsupervised loss at the beginning of training. It is observed that when the model is randomly initialized, the SRC loss is relatively low, which can be explained by that the relation matrix $R^l$ is also random and meaningless with no much difference under different perturbations. 
When the training starts, the SRC loss increases as the model starts learning discriminative features and produces meaningful relation matrix with differences under perturbations. With the consistency weight increasing, both the SRC loss and conventional consistency loss gradually decrease, leading to a higher classification performance with decreased cross entropy loss.  
}

\subsubsection{\textbf{Performance on other consistency-based frameworks}}
Our proposed sample-relation-consistency paradigm is a general mechanism to enhance the consistency-based semi-supervised learning framework. The choice of the baseline framework is flexible. 
Here, we incorporate the proposed SRC paradigm into two additional popular consistency-based frameworks, \ie, $\Pi$ model and Temporal Ensembling (TE). 
The performance of original semi-supervised learning framework and sample-relation-consistency enhanced version are listed in Table~\ref{tab:tab6}. 
As we can see, under 20\% labeled data setting, the SRC paradigm consistently improves the performance of $\Pi$ model and TE on AUC and F1 score metric, demonstrating that enforcing the relation consistency to better exploit the unlabeled data is a general and effective strategy under different semi-supervised frameworks. 

\begin{table}[t]
    \renewcommand\arraystretch{1.2}
	\centering
    \caption{Evaluation of SRC with different consistency-based frameworks on ISIC 2018 dataset.}
    \label{tab:tab6}
    \begin{tabular}{c|cc|cc}
        \hline
        \multirow{2}{*}{Method} & \multicolumn{2}{c|}{Percentage} &\multicolumn{2}{c}{Metrics} \\
        \cline{2-5}
                  &Labelled &Unlabelled    &AUC    &F1\\
        \hline
        $\Pi$~\cite{laine2017temporal} &20\%  &80\%          &92.24  &58.44 \\
        SRC-$\Pi$
                    &20\% &80\%   &92.90 &60.17 \\
        \hline
        TE~\cite{laine2017temporal} &20\%  &80\%          &92.70 &59.33\\
        SRC-TE   
                    &20\% &80\% &93.57 &59.70 \\
        \hline
    \end{tabular}
\end{table}
\begin{table}[t]
    \renewcommand\arraystretch{1.2}
	\centering
    \caption{Comparison of the feature consistency and sample relation consistency on ISIC2018 dataset.}
    \label{tab:feature_consistency}
    \scalebox{0.75}{
    \begin{tabular}{c|c|ccccc}
        \hline
        \multirow{2}{*}{Method} & \multicolumn{1}{c|}{hyper-parameter} &\multicolumn{5}{c}{Metrics} \\
        \cline{3-7}
                    & weight $\beta$    &AUC    &Sensitivity    &Specificity   &Accuracy    &F1    \\
        \hline
        MT
                    &0 &92.96 &69.75 &92.20 &92.48 &59.10\\
        \hline
        FC-MT 
                    &0.01      &93.04 &69.71 &92.72  &92.43 &58.21\\
        FC-MT 
                    &0.1      &93.13 &70.13 &92.35  &92.51 &59.81\\
        FC-MT 
                    &1.0       &92.32 &63.36 &92.01  &92.18 &55.04\\
        \hline
        SRC-MT (ours) 
                    &0.01      &93.08 &70.64 &92.49 &92.57 &59.68\\
        SRC-MT (ours) 
                    &0.1      &93.47 &71.01 &92.20  &92.63 &59.88\\
        SRC-MT (ours) 
                    &1.0        &93.58 &71.47&92.72&92.54&60.68\\
        \hline
    \end{tabular}
    }
\end{table}
\section{Discussions}
Automated disease classification from medical images is essential for assisting clinical diagnosis and treatment planning. 
For the sake of high performance, current deep learning approaches usually require a large amount of labeled data for network training. 
However, acquiring high-quality label for medical data is laborious and tedious work for the doctors. 
It has great potential to study semi-supervised medical image classification method to reduce the demand on labeled data by effectively exploiting the unlabeled data. 
Drawing spirit from the recent consistency enforcing methods in semi-supervised learning~\cite{tarvainin2017mean}, we adopt consistency mechanism to leverage the unlabeled data, which enforces the prediction consistency for the same input samples under different perturbations. A self-ensembling teacher model is further employed to produce more reliable consistency targets for the unlabeled data. 
More importantly, based on the observation that the doctors refer to previous analogous cases to make accurate diagnosis, we propose a novel sample-relation-consistency (SRC) paradigm to better exploit the unlabeled data by exploring the valuable relation information among different samples. 

The SRC paradigm requires the consistency of the intrinsic relation of data under different perturbations. Trained in this way, the network would be enhanced to learn more robust and discriminative representations for maintaining the semantic relation consistency, and thus helps to extract additional semantic information from unlabeled data. 
\revisemajor{
Theoretically, constraining the consistency of the semantic embedding $A^l$ could also guarantee the consistency of relation structure $R^l$. 
Here we explore how the model would preform if we directly enforce the consistency of $A^l$ under perturbations. We call this approach as feature consistency, and compare it with the proposed SRC paradigm.  Specifically, we applied these two consistency mechanisms respectively on the mean teacher framework, and evaluated their performance under different loss weight $\beta$. For the feature consistency strategy, we adopted the mean squared error as the regularization objective, which is the same as the SRC paradigm. As shown in Table~\ref{tab:feature_consistency}, the performance of feature consistency method (FC-MT) is inferior to the original mean teacher (MT) framework under the setting of $\beta = 1$, indicating that directly regularizing the consistency of latent features might be a too strict constraint that leads to negative effects. When relaxing this direct feature constraint by adjusting the loss weight $\beta$ to 0.1, the FC-MT is slightly superior to MT, whereas the improvements is marginal compared with our SRC-MT. When further decreasing $\beta$ to 0.01, both FC-MT and SRC-MT show limited advantage over MT, since the training loss would be dominated by the classification loss and the conventional consistency loss. Overall, the feature consistency approach would introduce too strict regularization into the framework. It is sensitive to the loss weight $\beta$ setting and can marginally improve the original MT framework. While our sample relation consistency paradigm could effectively exploit the relationship information among samples by adding regularization to the sample relation matrix $R^l$ not directly to feature $A^l$.
}

%
%

There are a variety of ways for estimating the relation (similarity) among samples. For example, the cosine similarity \revisemajor{was} utilized in face recognition to measure the similarity among feature vectors of different samples~\cite{wang2018cos}, and further extended for similarity measurement in lesions identification~\cite{luo2020angular}. In addition, Liu~\etal~\cite{liu2019knowledge} \revisemajor{constructed} an instance relation graph to model the relation structure of data. 
In this work, we formulate the similarity as the inner product of two vectors, inspired by the Gram matrix. As part of our future works, we would explore different approaches to estimate the similarity between samples and analyze their effect to the classification performance. 
\revisemajor{Besides, designing more effective perturbation scheme is able to improve the performance of consistency-based methods. There are some recent automatic data augmentation works~\cite{cubuk2019autoaugment} that automatically search the best transformations for a specific dataset. It is an interesting future work to explore how to utilize automatic data transformations to create better perturbation and maximize the benefits of consistency mechanism in semi-supervised learning.} Moreover, how to extend the relational consistency mechanism into semi-supervised segmentation problem, to effectively exploit the unlabeled data collected from other clinical centers~\cite{liu2020msnet} or other modalities~\cite{dou2020unpaired} is also an interesting direction for our future work.

\begin{figure}[t]
	\centering
	\revisemajor{
	\includegraphics[width=0.47\textwidth]{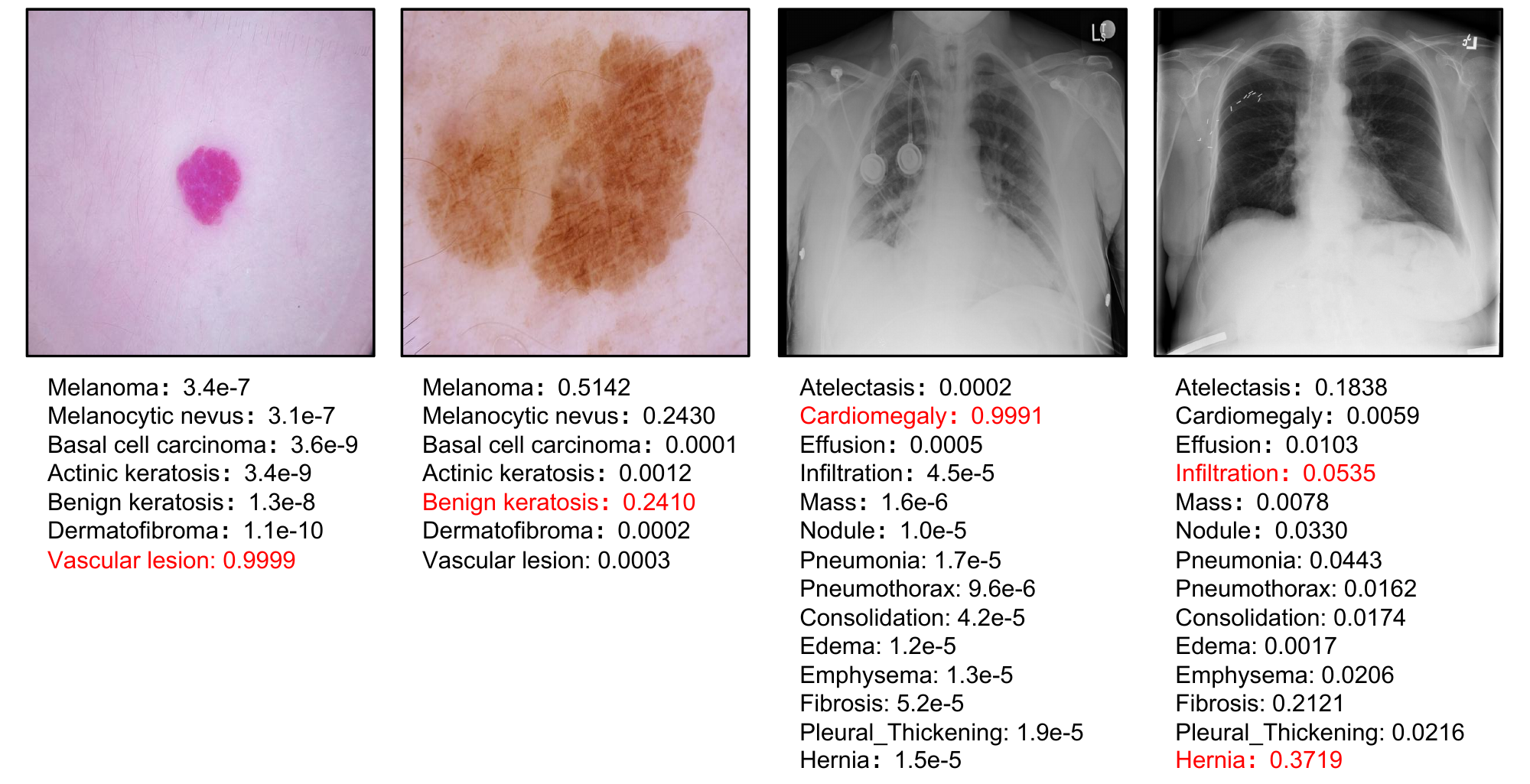}
     \caption{Typical examples with accurate and inaccurate predictions on the two applications. Classes with red color denote the ground truth label.}
	\label{fig:typical_example}
}
	\vspace{-3mm}
\end{figure}

~\revisemajor{We visualize some typical samples with the best and worst predictions of our model on the two applications in Fig.~\ref{fig:typical_example}, from which we see that the classification performance of our method still leaves room for improvement.}
In this work, we adopted the basic DenseNet not other complex network designs~\cite{gessert2019skin, yan2018weakly} as backbone since we focus on studying how to effectively exploit unlabeled data by leveraging the intrinsic relation of data. 
Integrating our proposed semi-supervised mechanism into other well-designed networks is one of our future works, for more accurate medical images classification in semi-supervised manner. 
In addition, even though our method achieves comparable AUC and Sensitivity with the upper-bound model on the skin lesion classification task, the improvement on Specificity is relatively limited. 
Based on this observation, in the future, we plan to study how to consistently improve discrimination ability on positive and negative samples, by exploring the class re-balancing techniques~\cite{cui2019class}.
%


In general, our proposed semi-supervised framework can be applied to different medical image classification applications, as we have demonstrated its effectiveness on both single-label and multi-label disease classification tasks. 
Meanwhile, the proposed SRC paradigm is of general feasibility to be incorporated to other semi-supervised frameworks, as illustrated in Table~\ref{tab:tab6}. 
It is worth noting that under fully supervised scenario, the SRC paradigm could also be utilized as another type of data augmentation scheme. 
Conventional data augmentation mechanism emphasizes the output to be invariant under different perturbations (\eg, rotation, translation and flipping). 
It is interesting to investigate whether the invariance of the semantic relation among different samples would help to alleviate the over-fitting in fully-supervised scenario.   

\section{Conclusion}
In this work, we study the semi-supervised medical image classification problem to reduce the human effort of labeling medical image data. 
We present a new semi-supervised self-ensembling framework by introducing a novel sample relation consistency (SRC) paradigm.
Our framework can better exploit unlabeled data by exploring the intrinsic relation between samples. 
Extensive experiments on two large-scale public benchmark datasets demonstrate the superiority of our method over the state-of-the-art semi-supervised learning methods on both single-label and multi-label medical image classification tasks. 
Moreover, the proposed SRC paradigm is a general strategy, which is feasible to be incorporated with other semi-supervised learning approaches.

\bibliographystyle{IEEEtran}
\small\bibliography{refs}

\begin{thebibliography}{10}
\providecommand{\url}[1]{#1}
\csname url@samestyle\endcsname
\providecommand{\newblock}{\relax}
\providecommand{\bibinfo}[2]{#2}
\providecommand{\BIBentrySTDinterwordspacing}{\spaceskip=0pt\relax}
\providecommand{\BIBentryALTinterwordstretchfactor}{4}
\providecommand{\BIBentryALTinterwordspacing}{\spaceskip=\fontdimen2\font plus
\BIBentryALTinterwordstretchfactor\fontdimen3\font minus
  \fontdimen4\font\relax}
\providecommand{\BIBforeignlanguage}[2]{{%
\expandafter\ifx\csname l@#1\endcsname\relax
\typeout{** WARNING: IEEEtran.bst: No hyphenation pattern has been}%
\typeout{** loaded for the language `#1'. Using the pattern for}%
\typeout{** the default language instead.}%
\else
\language=\csname l@#1\endcsname
\fi
#2}}
\providecommand{\BIBdecl}{\relax}
\BIBdecl

\bibitem{tschandl2018the}
P.~Tschandl, C.~Rosendahl, and H.~Kittler, ``The ham10000 dataset, a large
  collection of multi-source dermatoscopic images of common pigmented skin
  lesions.'' \emph{Scientific Data}, vol.~5, 2018.

\bibitem{rajpurkar2017chexnet}
P.~Rajpurkar, J.~Irvin, K.~Zhu, B.~Yang, H.~Mehta, T.~Duan, D.~Ding, A.~Bagul,
  C.~Langlotz, K.~Shpanskaya, M.~P. Lungren, and A.~Y. Ng, ``Chexnet:
  Radiologist-level pneumonia detection on chest x-rays with deep learning,''
  in \emph{https://arxiv.org/abs/1711.05225}, 2017.

\bibitem{laine2017temporal}
S.~Laine and T.~Aila, ``Temporal ensembling for semi-supervised learning,'' in
  \emph{International Conference on Learning Representations}, 2017.

\bibitem{tarvainin2017mean}
V.~H. Tarvainen, A., ``Mean teachers are better role models: Weight-averaged
  consistency targets improve semi-supervised deep learning results.'' in
  \emph{Adv. Neural Inf. Process. Syst.}, 2017.

\bibitem{cui2019semi}
W.~Cui, Y.~Liu, Y.~Li, M.~Guo, Y.~Li, X.~Li, T.~Wang, X.~Zeng, and C.~Ye,
  ``Semi-supervised brain lesion segmentation with an adapted mean teacher
  model,'' in \emph{Inf. Process. Med. Imaging}, 2019, pp. 554--565.

\bibitem{perone2018deep}
C.~S. Perone and J.~Cohen-Adad, ``Deep semi-supervised segmentation with
  weight-averaged consistency targets,'' in \emph{DLMIA}, 2018, pp. 12--19.

\bibitem{yu2019uncertainty}
L.~Yu, S.~Wang, X.~Li, C.-W. Fu, and P.-A. Heng, ``Uncertainty-aware
  self-ensembling model for semi-supervised 3d left atrium segmentation,'' in
  \emph{Medical Image Computing and Comput.-Assisted Intervention.}\hskip 1em
  plus 0.5em minus 0.4em\relax Springer, 2019, pp. 605--613.

\bibitem{liu2019knowledge}
Y.~Liu, J.~Cao, B.~Li, C.~Yuan, W.~Hu, Y.~Li, and Y.~Duan, ``Knowledge
  distillation via instance relationship graph,'' in \emph{Proc. IEEE Conf.
  Comput. Vis. Pattern Recognit.}, 2019.

\bibitem{Battaglia2018relational}
P.~W. Battaglia, J.~B. Hamrick, V.~Bapst, V.~Gonzalez, Alvaro
  Sanchez-and~Zambaldi, M.~Malinowski, A.~Tacchetti, D.~Raposo, A.~Santoro,
  R.~Faulkner, C.~Gulcehre, F.~Song, A.~Ballard, J.~Gilmer, G.~Dahl,
  A.~Vaswani, K.~Allen, C.~Nash, V.~Langston, C.~Dyer, N.~Heess, D.~Wierstra,
  P.~Kohli, M.~Botvinick, O.~Vinyals, Y.~Li, and R.~Pascanu, ``Relational
  inductive biases, deep learning, and graph networks,'' in
  \emph{https://arxiv.org/abs/1806.01261}, 2018.

\bibitem{avni2011xray}
U.~Avni, H.~Greenspan, E.~Konen, M.~Sharon, and J.~Goldberger, ``X-ray
  categorization and retrieval on the organ and pathology level, using
  patch-based visual words,'' \emph{IEEE Trans. Med. Imaging}, vol.~30, no.~3,
  pp. 733--746, 2011.

\bibitem{xie2019semi}
Y.~Xie, J.~Zhang, and Y.~Xia, ``Semi-supervised adversarial model for
  benign–malignant lung nodule classification on chest ct,'' \emph{Med. Image
  Anal.}, vol.~57, pp. 237--248, 2019.

\bibitem{miyato2019virtual}
T.~Miyato, S.~Maeda, M.~Koyama, and S.~Ishii, ``Virtual adversarial training: A
  regularization method for supervised and semi-supervised learning,''
  \emph{IEEE Trans. Pattern Anal. Mach. Intell.}, vol.~41, no.~8, pp.
  1979--1993, 2019.

\bibitem{cheplygina2018not}
V.~Cheplygina, M.~d. Bruijne, and J.~P.~W. Pluim, ``Not-so-supervised: a survey
  of semi-supervised, multi-instance, and transfer learning in medical image
  analysis,'' vol.~54, 2019, pp. 280--296.

\bibitem{bai2017semi}
W.~Bai, O.~Oktay, M.~Sinclair, H.~Suzuki, M.~Rajchl, G.~Tarroni, B.~Glocker,
  A.~King, P.~M. Matthews, and D.~Rueckert, ``Semi-supervised learning for
  network-based cardiac mr image segmentation,'' in \emph{Medical Image
  Computing and Comput.-Assisted Intervention.}\hskip 1em plus 0.5em minus
  0.4em\relax Springer, 2017, pp. 253--260.

\bibitem{zhang2017deep}
Y.~Zhang, L.~Yang, J.~Chen, M.~Fredericksen, D.~P. Hughes, and D.~Z. Chen,
  ``Deep adversarial networks for biomedical image segmentation utilizing
  unannotated images,'' in \emph{Medical Image Computing and Comput.-Assisted
  Intervention.}\hskip 1em plus 0.5em minus 0.4em\relax Springer, 2017, pp.
  408--416.

\bibitem{jin2019incorporating}
Y.~Jin, K.~Cheng, Q.~Dou, and P.-A. Heng, ``Incorporating temporal prior from
  motion flow for instrument segmentation in minimally invasive surgery
  video,'' in \emph{International Conference on Medical Image Computing and
  Computer-Assisted Intervention.}\hskip 1em plus 0.5em minus 0.4em\relax
  Springer, 2019, pp. 440--448.

\bibitem{gu2017semi}
L.~Gu, Y.~Zheng, R.~Bise, I.~Sato, N.~Imanishi, and S.~Aiso, ``Semi-supervised
  learning for biomedical image segmentation via forest oriented super
  pixels(voxels),'' in \emph{Medical Image Computing and Comput.-Assisted
  Intervention.}\hskip 1em plus 0.5em minus 0.4em\relax Springer, 2017, pp.
  702--710.

\bibitem{singh2011identifying}
S.~Singh, F.~Janoos, T.~Pécot, E.~Caserta, G.~Leone, J.~Rittscher, and
  R.~Machiraju, ``Identifying nuclear phenotypes using semi-supervised metric
  learning,'' in \emph{Inf. Process. Med. Imaging}, 2011, pp. 398--410.

\bibitem{bai2019self}
W.~Bai, C.~Chen, G.~Tarroni, J.~Duan, F.~Guitton, S.~E. Petersen, Y.~Guo, P.~M.
  Matthews, and D.~Rueckert, ``Self-supervised learning for cardiac mr image
  segmentation by anatomical position prediction,'' in \emph{International
  Conference on Medical Image Computing and Computer-Assisted
  Intervention}.\hskip 1em plus 0.5em minus 0.4em\relax Springer, 2019, pp.
  541--549.

\bibitem{xia2018cotraining}
Y.~Xia, F.~Liu, D.~Yang, J.~Cai, L.~Yu, Z.~Zhu, D.~Xu, A.~Yuille, and H.~Roth,
  ``3d semi-supervised learning with uncertainty-aware multi-view
  co-training,'' in \emph{IEEE Winter Conf. Appl. Comput. Vis.}, 2020.

\bibitem{sun2016computerized}
W.~Sun, T.~B. Tseng, J.~Zhang, and W.~Qian, ``Computerized breast cancer
  analysis system usingthree stage semi-supervised learning method,''
  \emph{Comput. Methods Programs Biomed.}, vol. 135, pp. 77--88, 2016.

\bibitem{baur2017semi}
C.~Baur, S.~Albarqouni, and N.~Navab, ``Semi-supervised deep learning for fully
  convolutional networks,'' in \emph{Medical Image Computing and
  Comput.-Assisted Intervention.}\hskip 1em plus 0.5em minus 0.4em\relax
  Springer, 2017, pp. 311--319.

\bibitem{chen2019discriminative}
J.~Chen, H.~Zhang, Y.~Zhang, S.~Zhao, R.~Mohiaddin, T.~Wong, D.~Firmin,
  G.~Yang, and J.~Keegan, ``Discriminative consistent domain generation for
  semi-supervised learning,'' in \emph{Medical Image Computing and
  Comput.-Assisted Intervention.}\hskip 1em plus 0.5em minus 0.4em\relax
  Springer, 2019, pp. 595--604.

\bibitem{chartsias2018factorised}
A.~Chartsias, T.~Joyce, G.~Papanastasiou, S.~Semple, M.~Williams, D.~Newby,
  R.~Dharmakumar, and S.~A. Tsaftaris, ``Factorised spatial representation
  learning: application in semi-supervised myocardial segmentation,'' in
  \emph{Medical Image Computing and Comput.-Assisted Intervention.}\hskip 1em
  plus 0.5em minus 0.4em\relax Springer, 2018, pp. 490--498.

\bibitem{nie2018asdnet}
D.~Nie, Y.~Gao, L.~Wang, and D.~Shen, ``Asdnet: Attention based semi-supervised
  deep networks for medical image segmentation,'' in \emph{Medical Image
  Computing and Comput.-Assisted Intervention.}\hskip 1em plus 0.5em minus
  0.4em\relax Springer, 2018, pp. 370--378.

\bibitem{aviles2019graph}
A.~I. Aviles-Rivero, N.~Papadakis, R.~Li, P.~Sellars, Q.~Fan, R.~T. Tan, and
  C.~Schönlieb, ``Graphx net − chest x-ray classification under extreme
  minimal supervision,'' in \emph{Medical Image Computing and Comput.-Assisted
  Intervention.}\hskip 1em plus 0.5em minus 0.4em\relax Springer, 2019, pp.
  504--512.

\bibitem{li2020transformation}
X.~Li, L.~Yu, H.~Chen, C.-W. Fu, L.~Xing, and P.-A. Heng,
  ``Transformation-consistent self-ensembling model for semi-supervised medical
  image segmentation,'' in \emph{IEEE T. NEUR. NET. LEAR.}, 2020.

\bibitem{dong2018unsupervised}
N.~Dong, M.~Kampffmeyer, X.~Liang, Z.~Wang, W.~Dai, and E.~P. Xing,
  ``Unsupervised domain adaptation for automatic estimation of cardiothoracic
  ratio,'' in \emph{Medical Image Computing and Comput.-Assisted
  Intervention.}\hskip 1em plus 0.5em minus 0.4em\relax Springer, 2018, pp.
  544--552.

\bibitem{andres2019retinal}
A.~Diaz-Pinto, A.~Colomer, V.~Naranjo, S.~Morales, Y.~Xu, and A.~F. Frangi,
  ``Retinal image synthesis and semi-supervised learning for glaucoma
  assessment.'' \emph{IEEE Trans. Med. Imaging}, vol.~38, no.~9, pp.
  2211--2218, 2019.

\bibitem{aviles2020when}
A.~I. Aviles-Rivero, N.~Papadakis, R.~Li, S.~M. Alsaleh, R.~T. Tan, and C.-B.
  Schonlieb, ``When labelled data hurts: Deep semi-supervised classification
  with the graph 1-laplacian,'' in \emph{https://arxiv.org/abs/1906.08635v3},
  2020.

\bibitem{li2018semi}
X.~Li, L.~Yu, H.~Chen, C.-W. Fu, and P.~A. Heng, ``Semi-supervised skin lesion
  segmentation via transformation consistent self-ensembling model,'' in
  \emph{The British Machine Vision Conference}, 2018.

\bibitem{su2019local}
H.~Su, S.~Xiaoshuang, J.~Cai, and L.~Yang, ``Local and global consistency
  regularized mean teacher for semi-supervised nuclei classification,'' in
  \emph{Medical Image Computing and Comput.-Assisted Intervention.}\hskip 1em
  plus 0.5em minus 0.4em\relax Springer, 2019, pp. 559--567.

\bibitem{ganster2001ganster}
H.~Ganster, A.~Pinz, R.~Röhrer, E.~Wildling, M.~Binder, and H.~Kittler,
  ``Automated melanoma recognition,'' \emph{IEEE Trans. Med. Imaging}, vol.~20,
  no.~3, pp. 233--239, 2001.

\bibitem{codella2015deep}
N.~C.~F. Codella, Q.~B. Nguyen, S.~Pankanti, D.~Gutman, B.~Helba, A.~Halpern,
  and J.~R. Smith, ``Deep learning ensembles for melanoma recognition in
  dermoscopy images,'' in \emph{IBM J. Res. Dev.}, vol.~61, no.~5, 2017, pp.
  1--15.

\bibitem{yu2016automated}
L.~Yu, H.~Chen, Q.~Dou, J.~Qin, and P.-A. Heng, ``Automated melanoma
  recognition in dermoscopy images via very deep residual networks,''
  \emph{IEEE Trans. Med. Imaging}, vol.~36, no.~4, pp. 994--1004, 2016.

\bibitem{xue2019robust}
C.~Xue, Q.~Dou, X.~Shi, H.~Chen, and P.~A. Heng, ``Robust learning at noisy
  labeled medical images: Applied to skin lesion classification,'' \emph{arXiv
  preprint arXiv:1901.07759}, 2019.

\bibitem{zhang2019attention}
J.~Zhang, Y.~Xie, Y.~Xia, and C.~Shen, ``Attention residual learning for skin
  lesion classification,'' \emph{IEEE Trans. Med. Imaging}, vol.~38, no.~9, pp.
  2092--2103, 2019.

\bibitem{shi2019active}
X.~Shi, Q.~Dou, C.~Xue, J.~Qin, H.~Chen, and P.-A. Heng, ``An active learning
  approach for reducing annotation cost in skin lesion analysis,'' in
  \emph{International Workshop on Machine Learning in Medical Imaging}.\hskip
  1em plus 0.5em minus 0.4em\relax Springer, 2019, pp. 628--636.

\bibitem{wang2017chest8}
X.~Wang, Y.~Peng, L.~Lu, Z.~Lu, M.~Bagheri, and R.~M. Summers, ``Chestx-ray8:
  Hospital-scale chest x-ray database and benchmarks on weakly-supervised
  classification and localization of common thorax diseases,'' in \emph{Proc.
  IEEE Conf. Comput. Vis. Pattern Recognit.}, 2017.

\bibitem{irvin2019chexpert}
J.~Irvin, P.~Rajpurkar, M.~Ko, Y.~Yu, S.~Ciurea-Ilcus, C.~Chute, H.~Marklund,
  B.~Haghgoo, R.~Ball, K.~Shpanskaya, J.~Seekins, D.~A. Mong, S.~S. Halabi,
  J.~K. Sandberg, R.~Jones, D.~B. Larson, C.~P. Langlotz, B.~N. Patel, M.~P.
  Lungren, and A.~Y. Ng, ``Chexpert: A large chest radiograph dataset with
  uncertainty labels and expert comparison,'' in \emph{The Thirty-Third AAAI
  inproceedings on Artificial Intelligence}, 2019.

\bibitem{yao2017learning}
L.~Yao, E.~Poblenz, D.~Dagunts, B.~Covington, D.~Bernard, and K.~Lyman,
  ``Learning to diagnose from scratch by exploiting dependencies among
  labels,'' in \emph{https://arxiv.org/abs/1710.10501}, 2017.

\bibitem{gatys2015gram}
L.~A. Gatys, A.~S. Ecker, and M.~Bethge, ``A neural algorithm of artistic
  style,'' in \emph{J. Vis.}, 2015.

\bibitem{huang2017dense}
G.~Huang, Z.~Liu, L.~v.~d. Maaten, and K.~Q. Weinberger, ``Densely connected
  convolutional networks,'' in \emph{Proc. IEEE Conf. Comput. Vis. Pattern
  Recognit.}, 2017.

\bibitem{codella2018skin}
N.~Codella, V.~Rotemberg, P.~Tschandl, M.~E. Celebi, S.~Dusza, D.~Gutman,
  B.~Helba, A.~Kalloo, K.~Liopyris, M.~Marchetti, H.~Kittler, and A.~Halpern,
  ``Skin lesion analysis toward melanoma detection 2018: A challenge hosted by
  the international skin imaging collaboration (isic),'' in
  \emph{https://arxiv.org/abs/1902.03368}, 2018.

\bibitem{russa2017imagenet}
O.~Russakovsky, J.~Deng, H.~Su, J.~Krause, S.~Satheesh, S.~Ma, Z.~Huang,
  A.~Karpathy, A.~Khosla, M.~Bernstein, A.~C. Berg, and F.-F. Li, ``Imagenet
  large scale visual recognition challenge,'' in
  \emph{https://arxiv.org/abs/1409.0575}, 2014.

\bibitem{kingma2015adam}
D.~P. Kingma and J.~Ba, ``Adam: A method for stochastic optimization,'' in
  \emph{International Conference on Learning Representations}, 2015.

\bibitem{gessert2019skin}
N.~Gessert, M.~Nielsen, M.~Shaikh, R.~Werner, and A.~Schlaefer, ``Skin lesion
  classification using ensembles of multi-resolution efficientnets with meta
  data,'' in \emph{https://arxiv.org/abs/1910.03910}, 2019.

\bibitem{yan2018weakly}
C.~Yan, J.~Yao, R.~Li, Z.~Xu, and J.~Huang, ``Weakly supervised deep learning
  for thoracic disease classification and localization on chest x-rays,'' in
  \emph{ACM-BCB}, 2018.

\bibitem{wang2018cos}
H.~Wang, Y.~Wang, Z.~Zhou, X.~Ji, D.~Gong, J.~Zhou, Z.~Li, and W.~Liu,
  ``Cosface: Large margin cosine loss for deep face recognition,'' \emph{Proc.
  IEEE Conf. Comput. Vis. Pattern Recognit.}, vol.~13, 2018.

\bibitem{luo2020angular}
L.~Luo, H.~Chen, X.~Wang, Q.~Dou, H.~Lin, J.~Zhou, G.~Li, and P.-A. Heng,
  ``Deep angular embedding and feature correlation attention for breast mri
  cancer analysis,'' in \emph{International Conference on Medical Image
  Computing and Computer-Assisted Intervention.}\hskip 1em plus 0.5em minus
  0.4em\relax Springer, 2019, pp. 504--512.

\bibitem{cubuk2019autoaugment}
E.~D. Cubuk, B.~Zoph, D.~Mane, V.~Vasudevan, and Q.~V. Le, ``Autoaugment:
  Learning augmentation strategies from data,'' in \emph{Proc. IEEE Conf.
  Comput. Vis. Pattern Recognit.}, 2019.

\bibitem{liu2020msnet}
Q.~Liu, Q.~Dou, L.~Yu, and P.~A. Heng, ``Ms-net: Multi-site network for
  improving prostate segmentation with heterogeneous mri data,'' in \emph{IEEE
  Trans. Med. Imaging}, 2020.

\bibitem{dou2020unpaired}
Q.~Dou, Q.~Liu, P.~A. Heng, and B.~Glocker, ``Unpaired multi-modal segmentation
  via knowledge distillation,'' in \emph{IEEE Trans. Med. Imaging}, 2020.

\bibitem{cui2019class}
Y.~Cui, M.~Jia, T.-Y. Lin, Y.~Song, and S.~Belongie, ``Class-balanced loss
  based on effective number of samples,'' in \emph{Proc. IEEE Conf. Comput.
  Vis. Pattern Recognit.}, 2019.

\end{thebibliography}

\end{document}